\documentclass[lettersize,journal]{IEEEtran}
\usepackage{amsmath,amsfonts}
\usepackage{algorithmic}
\usepackage{algorithm}
\usepackage{array}
\usepackage[caption=false,font=normalsize,labelfont=sf,textfont=sf]{subfig}
\usepackage{textcomp}
\usepackage{stfloats}
\usepackage{url}
\usepackage{verbatim}
\usepackage{graphicx}
\usepackage{cite}
\usepackage{multirow}
\usepackage{epsfig}
\usepackage{amsmath}
\usepackage{amssymb}
\usepackage{enumerate}
\usepackage{soul}
\usepackage{url}
\usepackage{color}
\usepackage[utf8]{inputenc}
\usepackage{capt-of}
\newcommand{\etal}{\textit{et al.}}
\newcommand{\eat}[1]{}
\begin{document}

\title{KTN: Knowledge Transfer Network for Learning Multi-person 2D-3D Correspondences}

\author{Xuanhan Wang,
	Lianli Gao,~\IEEEmembership{Member,~IEEE,}
	Yixuan Zhou,\\
	Jingkuan Song,~\IEEEmembership{Senior Member,~IEEE,}
	and Meng Wang,~\IEEEmembership{Fellow,~IEEE}
}

\markboth{Journal of \LaTeX\ Class Files,~Vol.~14, No.~8, August~2021}%
{Shell \MakeLowercase{\textit{et al.}}: A Sample Article Using IEEEtran.cls for IEEE Journals}


\maketitle

\begin{abstract}
Human densepose estimation, aiming at establishing dense correspondences between 2D pixels of human body and 3D human body template, is a key technique in enabling machines to have an understanding of people in images. It still poses several challenges due to practical scenarios where real-world scenes are complex and only partial annotations are available, leading to incompelete or false estimations. In this work, we present a novel framework to detect the densepose of multiple people in an image. The proposed method, which we refer to Knowledge Transfer Network (KTN), tackles two main problems: 1) how to refine image representation for alleviating incomplete estimations, and 2) how to reduce false estimation caused by the low-quality training labels (i.e., limited annotations and class-imbalance labels). Unlike existing works directly propagating the pyramidal features of regions for densepose estimation, the KTN uses a refinement of pyramidal representation, where it simultaneously maintains feature resolution and suppresses background pixels, and this strategy results in a substantial increase in accuracy. Moreover, the KTN enhances the ability of 3D based body parsing with external knowledges, where it casts 2D based body parsers trained from sufficient annotations as a 3D based body parser through a structural body knowledge graph. In this way, it significantly reduces the adverse effects caused by the low-quality annotations. The effectiveness of KTN is demonstrated by its superior performance to the state-of-the-art methods on DensePose-COCO dataset. Extensive ablation studies and experimental results on representative tasks (e.g., human body segmentation, human part segmentation and keypoints detection) and two popular densepose estimation pipelines (i.e., RCNN and fully-convolutional frameworks), further indicate the generalizability of the proposed method. 
\end{abstract}


\section{Introduction}
\label{sec:intro}
Human densepose estimation has been a fundamental topic in computer vision community for recent years. It aims to establish correspondences between the 2D human image pixels and the 3D human body template, which is a fundamental and critical task in the field of human-centric understanding. Inspired by successful works on human-centric recognition tasks, involving pedestrian detection \cite{humandet:occlusion-aware,humandet:repulsion,DBLP:journals/tcsv/JiYSSZ21}, human attribute recognition \cite{humanparsing:instance-level,humanparsing:understanding,densepose:parsingrcnn,DBLP:journals/tcsv/LinWLC0S21} and human keypoints detection \cite{pose_xiao2018simple,pose_est:mul_p,pose_li2018crowdpose,ins_seg:mask_rcnn,pose_SunXLWang2019,DBLP:journals/tcsv/JiaoYX21}, the two-stage strategy has become the dominant solution for current denspose estimation methods, namely, detecting people, extracting region features therefrom, and then jointly recognizing the body surface and detecting surface-specific UV coordinates. Based on this paradigm, high-quality densepose estimation depends on two key aspects: (1) whether the extracted region features are representative enough to make dense predictions, and (2) whether the densepose estimator is optimal for fully densepose parsing. Similar to other human-centric tasks, a well-established and commonly-used annotated dataset is critical for training a two-stage based densepose estimator. To achieve this, G{\"{u}}ler \etal~\cite{densepose:dphuman} collected a large-scale dataset (i.e., DensePose-COCO) and tried to make the densepose estimation possible. However, our study reveals that the typical densepose estimation pipeline suffers from several challenges, resulting in inaccurate estimation. In particular, there are two main issues that exist in current approaches, as analyzed below:
\begin{figure}[ht]
	\centering
	\setlength{\abovecaptionskip}{0pt}%
	\setlength{\belowcaptionskip}{0.cm}%
	\begin{center}
		\includegraphics[width=1.0\linewidth]{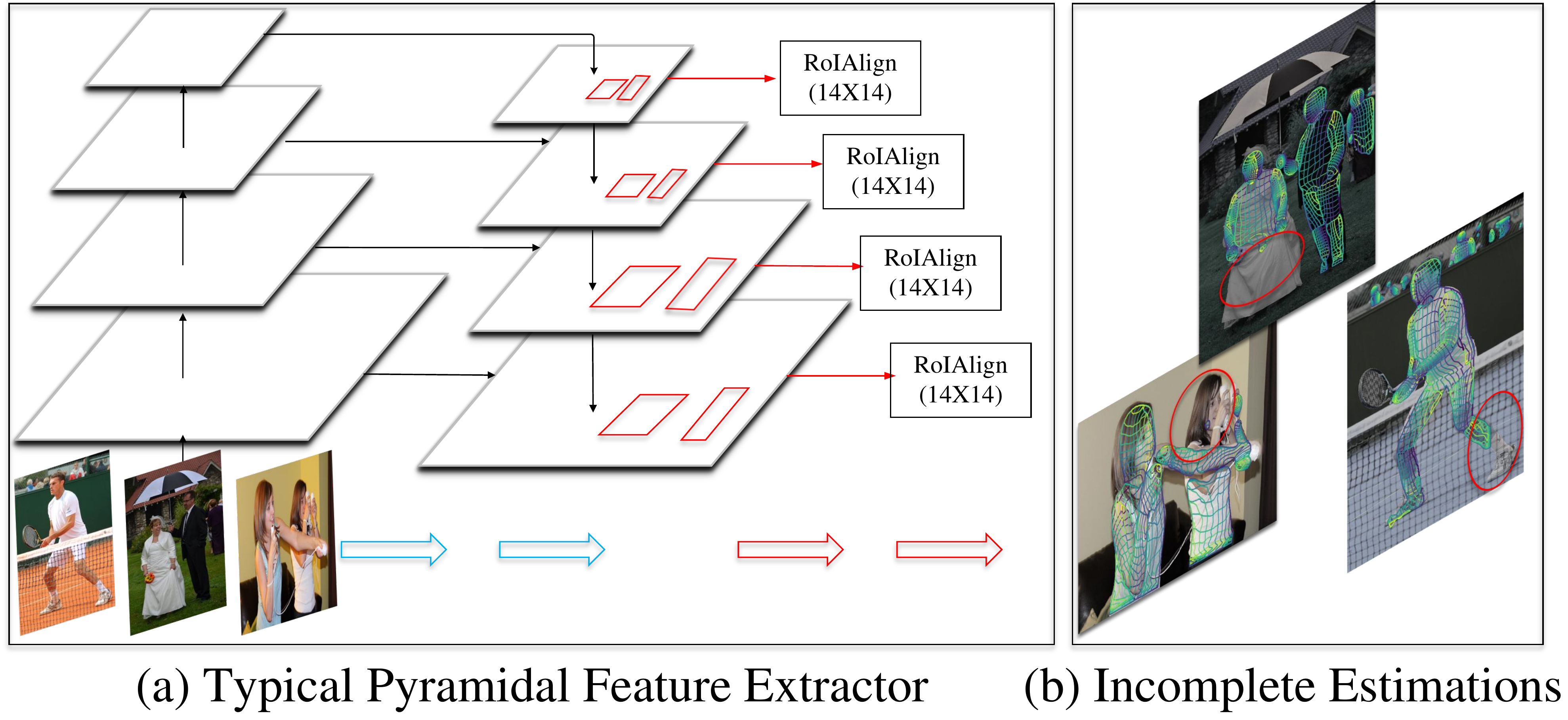}
	\end{center}
	
	\caption{Pyramidal representations used for densepose estimation: Region features derived from pyramidal image-level representations (a), which are either semantically poor or spatially coarse, lead to (b) incomplete estimation results.}
	\label{fig:issues}
\end{figure}

\noindent\textbf{Pyramidal representation:} Fig.~\ref{fig:issues}(a) indicates pyramidal based network topology, which is commonly used in current densepose estimation methods \cite{densepose:amanet,densepose:dphuman,densepose:parsingrcnn}. In particular, the region representation extracted from pyramidal embeddings may lose a lot of object details, leading to incomplete densepose estimation (Fig.~\ref{fig:issues}(b)). Specifically, there are two main reasons behind this: First, pyramid-like backbone network decreases resolution size as the network depth becomes deeper, lacking a unified representation that is semantically richer and spatially more precise as well. Previous work \cite{vis_cnn:cnn} has demonstrated that representations in shallow layers are spatially precise but semantically poor, while the representations in deep layers are strongly responded to high-level semantics but spatially sparse. Besides, a single-scale convolution operation used in all computation units lacks the diversity of the visual receptive field and cannot well handle the ``multi-scale'' setting in real-world scenes. Secondly, pyramid-like backbone network encodes visual information from the input image without constraining pixels of the background area. The region representation points extracted from pyramidal features may be noisy, and sometimes the activations of background area are larger than those corresponding to the foreground instances, as presented in \cite{ins_seg:shapeaware}. Therefore, directly utilizing typical pyramidal representation limits the overall performance on multi-person densepose estimation.

\begin{figure}[ht]
	\centering
	\setlength{\abovecaptionskip}{0pt}%
	\setlength{\belowcaptionskip}{0.cm}%
	\begin{center}
		\includegraphics[width=1.0\linewidth]{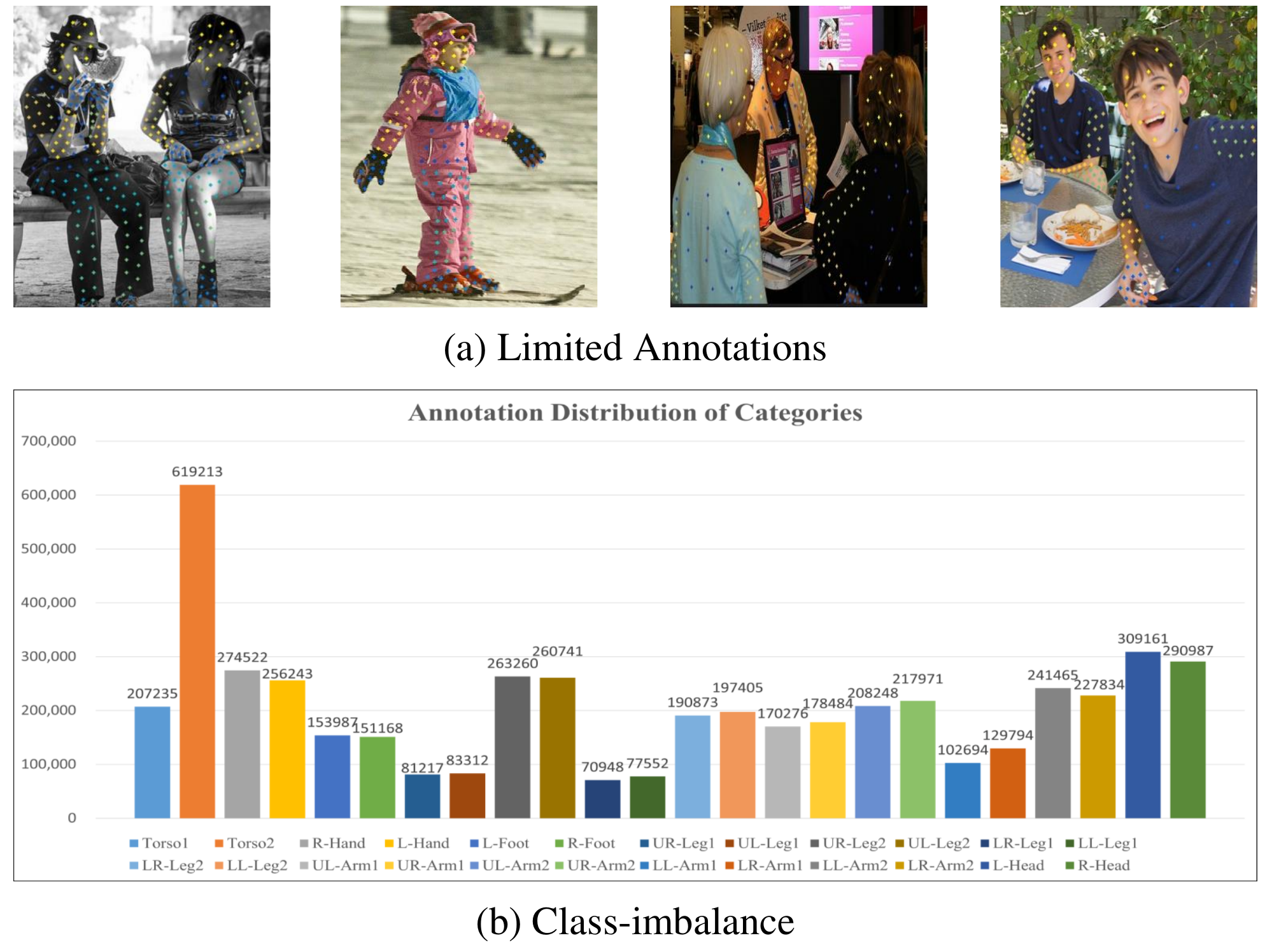}
	\end{center}
	
	\caption{Low-quality training labels in DensePose-COCO dataset: (a) Limited Annotations: Sparse pixel points (196 at most) are annotated with densepose labels for each person instance in an image. And (b) Class-imbalance problem: Annotation distribution of 3D surface categories in Densepose-COCO dataset presents the characteristic of imbalance.}
	\label{fig:issues2}
\end{figure}
\noindent\textbf{Low-quality Training Labels:} High-quality training dataset is crucial for the instantiation of a robust densepose estimator. However, the current dataset presents the characteristic of low quality due to heavy labor costs, particularly involving the issue of \textit{limited annotation} and the \textit{class-imbalance} problem.

In terms of the former issue, each person instance in the current available dataset (i.e., DensePose-COCO) has only sparse pixels (196 at most) with ground-truth 2D-3D correspondences, as depicted in Fig.~\ref{fig:issues2}(a). Furthermore, data diversity is particularly crucial for training an estimation model under the setting of supervised learning, and it is difficult to learn precise 2D-3D correspondence from such insufficient annotation resources. Therefore, the densepose estimation model that is trained under an limited supervision, would predict numerous false positives.

As for the latter issue, we observe that the number of labeled points for limbs such as arm or leg is much less than that for the torso or head, as shown in Fig.~\ref{fig:issues2}(b). This reveals that the training process of densepose estimation is significantly imbalanced. Furthermore, learning precise correspondences for limbs is much harder than that for torso and head, since limbs are more flexible than torso or head. Therefore, the class-imbalance problem, in which the rare categories are overwhelmed by the majority ones, will exacerbate the difficulty of learning 2D-3D correspondences at limbs.

The aforementioned challenges motivate us to answer two questions: 1) how to refine image representation for alleviating incomplete estimations, and 2) how to reduce false estimation caused by the low-quality annotations (i.e., limited annotations and class-imbalance problem). 
In terms of the former issue, a key observation is that a unified representation is needed, and feature points of backgrounds are noisy and confusing, which should be suppressed.
As for the latter one, we enrich the information contained in insufficient annotations by taking advantage of commonsense knowledge and annotation resources of 2D based human profiles (e.g., location, shape and keypoints).
In particular, humans are able to learn how to estimate densepose from such complex situations, with the help of commonsense knowledge. For example, humans will first search the commonsense knowledge about the structure of human body. When they identify the surface of upper arms in Fig.~\ref{fig:issues}(b), they will reason where upper arms are if they know the locations of shoulders, since the arm and the shoulder are strongly related according to the commonsense knowledge.
Furthermore, 2D based human profile annotations are much richer than the counterpart of 3D surface categories. However, the importance of commonsense knowledge addressing densepose estimation is significantly ignored by previous works, especially the knowledge about the relationship between the 2D based visual profile and 3D human body template.

Motivated by the above analysis, we develop a novel densepose estimation framework based on the two-stage pipeline, called \textbf{Knowledge Transfer Network} (KTN). Different from existing works that directly propagate the pyramidal base feature of regions, we enhance their representation power by a well-designed {\normalsize\bf{multi-instance decoder (MID)}}, which provides a unified image representation and suppresses the activations of backgrounds. To better handle the issue of limited annotations and imbalanced labels, we introduce a plug-and-play {\normalsize\bf{knowledge transfer machine (KTM)}}, which facilitates densepose estimation by utilizing the external commonsense knowledge.
Current densepose estimation methods either from RCNN based framework or fully-convolutional framework can be promoted in terms of the accuracy of human densepose estimation after equipping with KTM.

In summary, our work has three main contributions:
\begin{enumerate}[(1)]
	\item We propose an effective and end-to-end densepose estimation method named \textit{knowledge transfer network} (KTN), which addresses the issue of pyramidal representation and handles the problem of learning 2D-3D correspondences from insufficient and imbalanced labels.
	\item We are the first to introduce the knowledge to densepose estimation task. Our {knowledge transfer machine} can be easily embedded to any densepose estimation approaches either from RCNN based framework or fully-convolutional framework.
	\item Extensive experiments show that KTN can achieve significant densepose estimation performance. Code and models are also released for research purpose\footnote{\url{https://github.com/stoa-xh91/HumanDensePose}}.
\end{enumerate}

\section{Related Work}
\label{sec:relwork}
In this section, we first revisit recent advances in human densepose estimation. Then we review the progress in the field of image feature representation, few-shot recognition and class-imbalance problems.

\noindent\textbf{DensePose Estimation:}
Current state-of-the-art approaches of densepose estimation follow the two-stage pipeline for better performance. In particular, the Densepose-RCNN \cite{densepose:dphuman}, which follows the idea of Mask R-CNN \cite{ins_seg:mask_rcnn} and extends a densepose head in the typical two-stage detector, provides promising results on the densepose benchmark and become the first two-stage densepose estimation system. Furthermore, Parsing-RCNN \cite{densepose:parsingrcnn} introduces a human parsing approach to address the issue of instance details, where they enhance region feature representation by multi-scale dilated convolutions. AMANet \cite{densepose:amanet} further improves densepose performance by addressing the issue of scale diversity, where a multi-path technique is proposed to perceive multi-scale visual information. Although these different approaches vary in network topology, they all share a characteristic: they focus on the extraction of region-based densepose features and ignore the deficiency of pyramidal topology. In a different line of densepose estimation, recent works \cite{densepose:simpose,densepose:slimdp,densepose:uncertainty} focus on the issue of limited annotations. In particular, the Slim DP \cite{densepose:slimdp} aims at efficient densepose data collection and reduce the annotation effort by jointly utilizing video motion and sparse annotations. Furthermore, an uncertainty-aware training pipeline is proposed in \cite{densepose:uncertainty}, where it determines whether the estimation from the densepose model is confident or not, and relieves the model bias derived from limited training data. In \cite{densepose:simpose}, a domain-adaptation technique is proposed for training densepose system, where the densepose estimation model is jointly trained on simulated people and natural images of people. With the general success of these works, we make a further step towards simultaneously addressing the deficiency of feature representation and the issue of low-quality labels (i.e., sparse annotations and class-imbalance based recognition). 

\noindent\textbf{Image Feature Representation:}
Most developed networks for position sensitive tasks, such as object detection \cite{obj_det:FPN,obj_det:trident,obj_det:faster}, instance segmentation \cite{ins_seg:mask_rcnn,ins_seg:masklab,ins_seg:pan}, keypoints detection \cite{pose_xiao2018simple,pose_li2018crowdpose} and densepose estimation \cite{densepose:dphuman,densepose:amanet,densepose:parsingrcnn}, follow the design rule of pyramidal topological structure as depicted in Fig.~\ref{fig:issues}(a): gradually reduce the spatial size of the feature embeddings, and connect them from high resolution to low resolution, which provides various image representations for downstream tasks. However, typical pyramidal structure lacks the unified representation that is semantically rich and spatially precise. To achieve this, recent works \cite{pose_SunXLWang2019,img_seg:gridnet,img_seg:ms_densenet} devote to network design and follow a new design rule: maintaining high-resolution representations. Specifically, it maintains four-resolution embeddings in each layer, and connects them from low resolution to high resolution for final unified representation. However, such strategy maintains high-resolution representation front-to-end and brings a large computation burden. Instead of maintaining high-resolution representations front-to-end, we follow the design rule of pyramidal structure but apply a light module for unified image representation generation.

\noindent\textbf{Few-Shot Recognition:}
Few-shot learning is a training technique to understand new concepts with limited annotated examples. Early works \cite{fewshot:prototypical,fewshot:matching,zeroshot:wang2018zero-shot,fewshot:marino2017the,fewshot:jing2020self} use metric-learning approaches to learn feature representations of new concepts with limited annotation examples. In general, they try to preserve the class discriminative properties, where features of the semantically similar objects are closer than features of irrelevant objects. 
In a different line of few-shot learning, most recent works \cite{fewshot:dynamic,fewshot:model-agnostic,fewshot:sun2020meta} try to make use of the distillation of classifier parameters, which are trained from sufficient annotated examples, to help the few-shot classifier.
Different from them, we provide a commonsense knowledge aware distillation scheme, facilitating better densepose estimation. Besides, how to build an effective few-shot learning pipeline for densepose estimation, is still not explored. 

\noindent\textbf{Class-imbalance based Recognition:}
Long-tailed distribution is the fundamental property of class-imbalance problem, and it has been explored for various vision tasks. In general, current class-imbalance based recognition approaches can be categorized into three types: 1) re-sampling based methods \cite{longtail:exploring,longtail:relay}; 2) re-weighting based methods \cite{longtail:learning,longtail:modeltail,longtail:gradient,longtail:focal}; and 3) feature-based approaches \cite{longtail:yin2019feature,longtail:large-scale}. In particular, the re-sampling technique is one of the commonly used methods, which samples more training data from the rare classes. Similarly, re-weighting based methods try to assign different weights to the different classes for the balance purpose. As analyzed in \cite{imbalance_equaloss}, both of them are sensitive to the setting of the sample rate or the way of weight assignment, which makes the model harder to train. In a different line of these works, recent approaches \cite{longtail:yin2019feature,longtail:large-scale} try to build feature space of rare classes by transferring the features of regular classes that have sufficient annotations. Moreover, Reasoning-RCNN \cite{longtail:reasoning-rcnn} makes use of the external knowledge to augment the feature representations of rare classes and Forest R-CNN \cite{longtail:forset-rcnn} preserves more positive samples for rare categories by setting higher threshold during proposal generation. Instead of directly enhancing feature space or data space, we handle class-imbalance problem that exists in densepose estimation pipeline by improving the densepose estimator, where it recognizes body surfaces with the support of commonsense knowledge.

\noindent\textbf{Our approach:} Our knowledge transfer network (KTN) handles the deficiency of feature representation by introducing a multi-instance decoder (MID). It provides unified image representations that are semantically rich and spatially precise. In addition, the KTN handles the issue of low-quality densepose labels by enhancing densepose estimator with the help of external knowledge. Our knowledge transfer network can be viewed as an early attempt to explore external knowledge in the area of human densepose estimation.

This paper represents a very substantial extension of our conference version \cite{densepose:ktn}. Compared with \cite{densepose:ktn}, there are four main technical novelties:
(1) We extend the method proposed in \cite{densepose:ktn}, termed as KTNetV1, to a new version KTNetV2, which further explores the MID and the KTM. 
(2) We extend the MID module by introducing two resizing operations for enhancing feature representations.
(3) We build multiple 2D-3D transfer paths, which transfers multi-task priors (e.g., locations, body parts, keypoints) to the surface classifier for better densepose estimation.
(4) We show the superiority of proposed KTNet and present the generalizability of proposed techniques by evaluating them in a range of vision tasks or across various estimation pipelines.

\begin{figure*}[t]
	\begin{center}
		\includegraphics[width=0.9\linewidth]{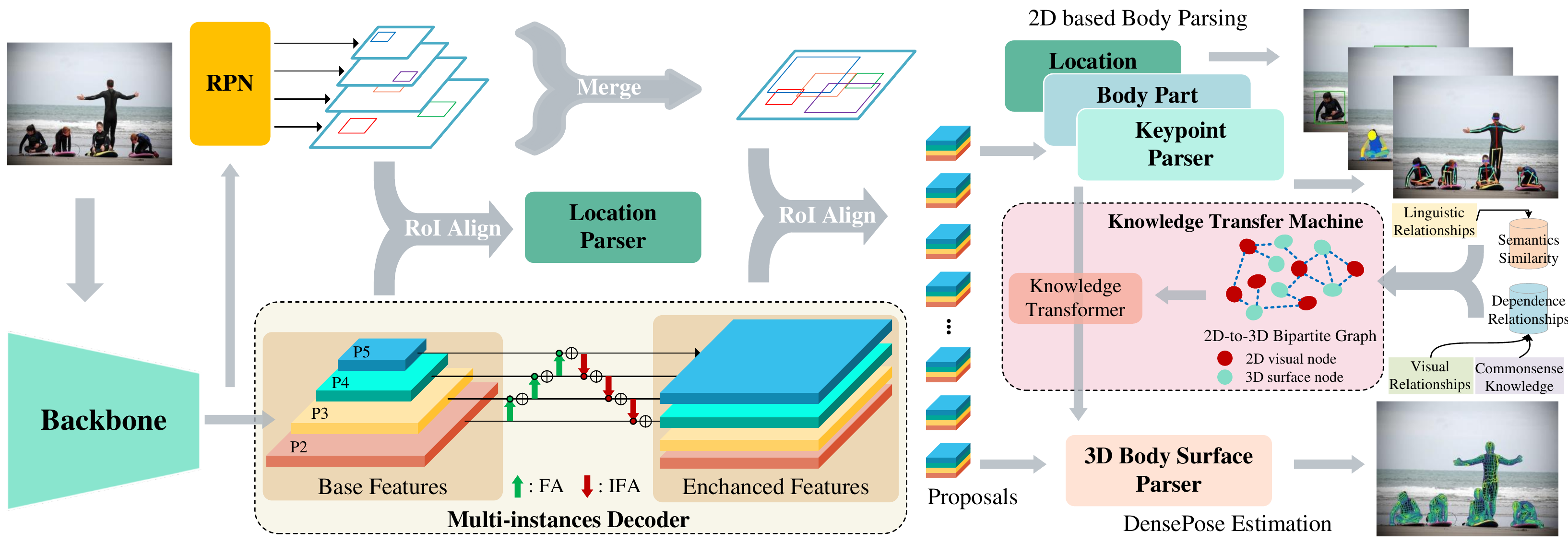}
	\end{center}
	\caption{The framework of KTN. The pyramidal base features are enhanced by the multi-instances decoder (MID), which preserves the instance details while suppresses the activations of backgrounds. The proposed knowledge transfer machine (KTM) transfers parameters of 2D based body parsers to a 3D surface classifier via commonsense knowledge, which alleviates adverse effects caused by low-quality labels.}
	\label{fig:framework}
	\hspace{-1.em}
\end{figure*}
\section{Knowledge Transfer Network}
\subsection{Approach Overview}
Fig.~\ref{fig:framework} illustrates the framework of our proposed Knowledge Transfer Network (KTN). Specifically, it consists of three main components. First, a base visual encoder is applied to extract base features from an image and outputs human detection results. In particular, we adopt two-stage object detection method based on the FPN as our base visual encoder, which is widely used in many related works \cite{densepose:amanet,densepose:dphuman,densepose:parsingrcnn}. Then, the base visual features are enhanced by proposed multi-instances decoder (MID), aiming to suppress the interference of background. Finally, a 3D body surface parser with the proposed Knowledge Transfer Machine (KTM) is used for instance-level human densepose estimation. In the following sections, we present details of each component.

\subsection{Multi-instances Decoder}
\begin{figure}[ht]
	\begin{center}
		\includegraphics[width=1.0\linewidth]{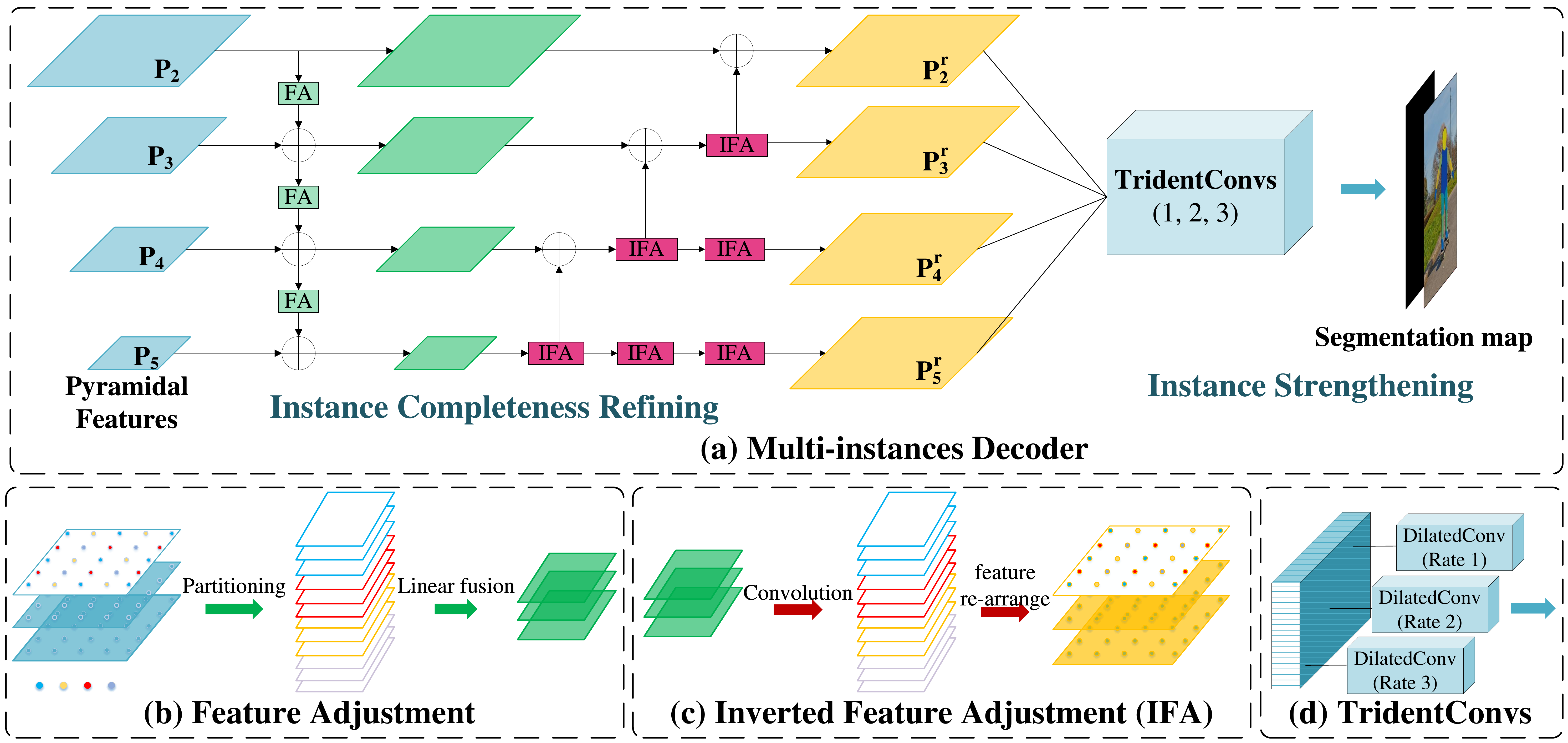}
	\end{center}
	\caption{The illustration of the Multi-instances Decoder (MID). Object details of multi-instances with diverse scales can be preserved in ICR step, while instances-of-interest are highlighted in IS step.}
	\label{fig:MID}
\end{figure}

To generate image-level feature representations that are semantically rich and spatially precise, we propose the Multi-instances Decoder (MID), as shown in Fig.~\ref{fig:MID}. Specifically, the MID consists of two steps: 1) Instance Completeness Refining (ICR) and 2) Instance Strengthening (IS). 

Before describing the above two processes, we first describe two rearranging operations, respectively named feature adjustment (FA) and inverted feature adjustment (IFA). Assuming a tensor $T \in R^{C \times H \times W}$, FA (Fig.~\ref{fig:MID}(b)) partitions it into four sub-tensors with the size of $\{C \times \frac{H}{2} \times \frac{W}{2}\}$ by recurrently sampling feature points at odd or even positions, outputting a resolution compressed yet spatially complete tensor $ \hat{T} \in R^{4C \times \frac{H}{2} \times \frac{W}{2}}$. In contrast to FA, IFA (Fig.~\ref{fig:MID}(c)) upsamples the tensor $T$ to a high resolution $2H \times 2W$ by predicting four feature maps with convolution operations, each of which is further distributed at odd or even positions in high resolution feature maps.

\subsubsection{Instance Completeness Refining} In general, pyramidal features involve four image-level feature representations, where feature resolution decreases by a factor of 2 as the level increases. Formally, we denote the pyramidal features as $\{P_2, P_3, P_4, P_5\}$ for simplicity. It is worth noting that the features at high levels (i.e., $P_4$, $P_5$) are spatially coarser but semantically stronger than that of features at low levels. However, the lowest level features (i.e., $P_2$) have the highest resolution and preserve the most complete instance details than that of high-level features. To break through these limits of typical pyramidal representations, we need to enhance semantic information at low-level features and preserve instance details at high-level features. To achieve this, we propose an instance completeness refining (ICR), which involves above two rearranging operations. Specifically, we hierarchically perform element-wise fusion (e.g., element-wise summation) between features at the high level (i.e., $P_i$, $i=3,4,5$) and lower level features (i.e., $P_{i-1}$, $i=3,4,5$), where each low-level feature is adjusted by the FA operation before the element-wise fusion. Then, fused features at the higher level are hierarchically added into lower-level features after they are dilated back to $2\times$ resolution by the IFA operation, resulting in spatially identical but refined features termed as $\{P_2^r, P_3^r, P_4^r, P_5^r\}$. Ultimately, we concatenate all levels of refined features and build trident convolutions \cite{obj_det:trident} to perceive multi-scale instances, resulting in a unified feature representation $P^c$ that is semantically rich and spatially precise.

\subsubsection{Instance Strengthening} 
In the IS step, our goal is to highlight informative yet instance-aware areas of image-level feature representation $P^c$, while suppress noisy background areas. To achieve this, we regress segmentation maps $M$ for instances-of-interest from the representation $P^c$, and enforce a pixel-wise classification loss on the segmentation maps during training. With such pixel-wise constrain, the feature areas of background are suppressed and pixels at the instance-aware area are activated, which is demonstrated in Sec.~\ref{sec.exp_mid_comp_details}.

\subsection{Knowledge Transfer Machine}
\begin{figure}[t]
	\begin{center}
		\includegraphics[width=1.0\linewidth]{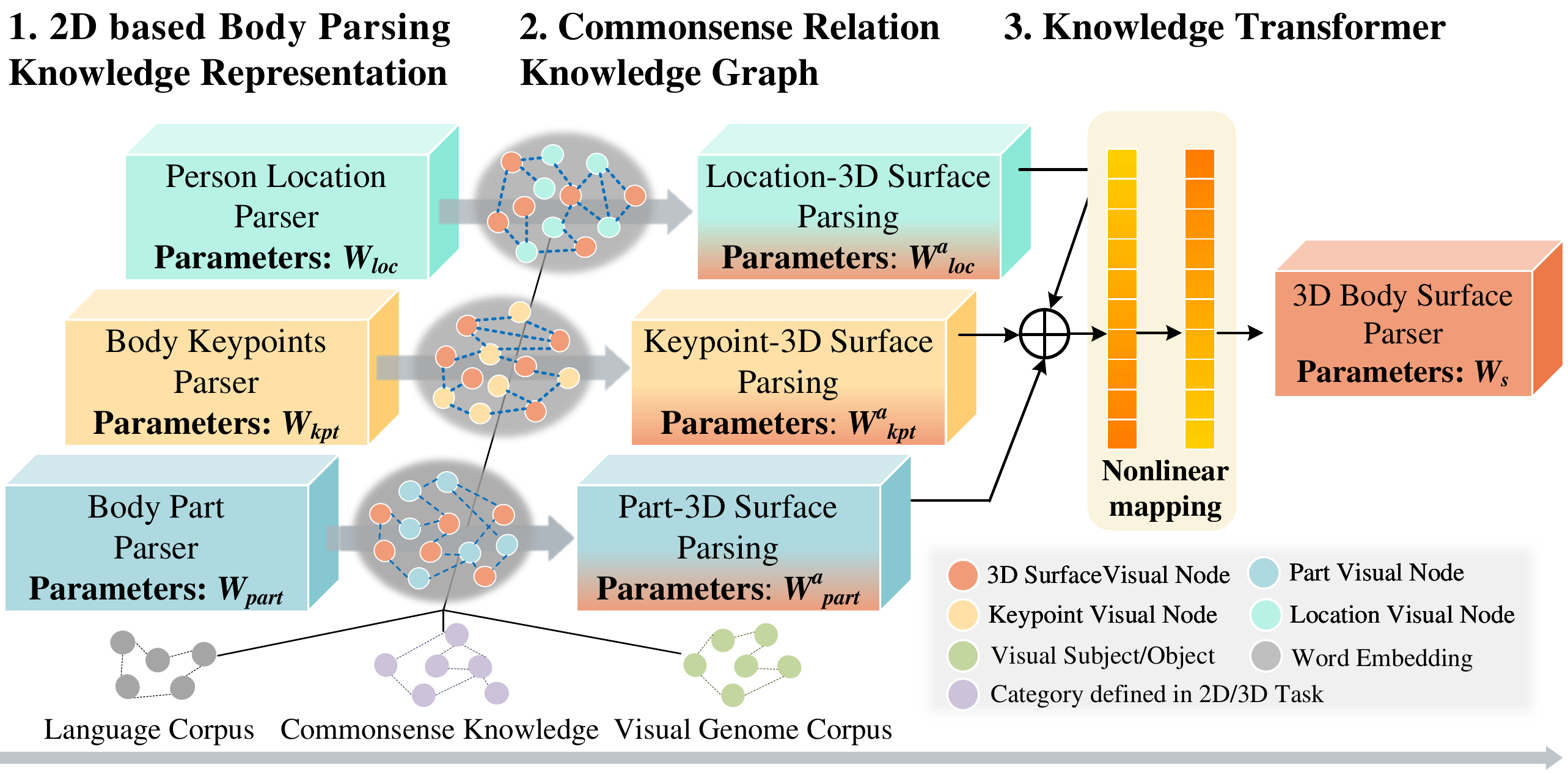}
	\end{center}
	\caption{The illustration of the Knowledge Transform Machine (KTM). With the help of commonsense relation knowledge graph, the parameters of 3D surface parser are generated via transferring parameters of 2D based body parsers.}
	\label{fig:KTM}
	\vspace{-1.5em}
\end{figure}

In this section, we introduce Knowledge Transfer Machine (KTM) to build a general model with commonsense knowledges to facilitate instance-level densepose estimation in the wild. In detail, we first construct the representation of 2D based body parsing knowledge by collecting parameters of 2D based body parsers, involving a person location parser, a body part parser, and a body keypoints parser. Then a 2D-to-3D directed bipartite graph $\mathit{G}: \mathit{G} = < C_{n}, C_{s}, \mathit{E} > $ is defined and used to associate most relevant parsing parameters for knowledge transferring, where $n\in \{loc, part, kpt\}$ denotes three 2D based tasks (i.e., person detection, part segmentation and keypoint estimation, respectively) and $s$ is the 3D based task (i.e., densepose estimation). $|C_n|$ denotes the total number of 2D visual nodes for task $n$ and $|C_s|$ is the total number of 3D surface nodes. In this work, each node represents a specific category defined in the corresponding task. For example, there are 17 categories (i.e., $|C_{kpt}|=17$) are defined in keypoint estimation task, 14 classes (i.e., $|C_{part}|=14$) in the part segmentation task and two classes (i.e., $|C_{loc}|=2$) in the person detection task. $E$ is the set of edges, each of which connects a 2D visual node to one of 3D surface node, and encodes the relation knowledge between two nodes. Finally, a knowledge transformer is adopted to cast the parameters of 2D based body parsers as the parameters of a 3D body surface parser. In summary, the proposed Knowledge Transfer Machine (KTM) consists of three main modules as shown in Fig.~\ref{fig:KTM}: 1) 2D based Body Parsing Knowledge Representation; 2) Commonsense Relation Knowledge Graph; and 3) Knowledge Transformer. Furthermore, the KTM can be applied in any RCNN based methods \cite{densepose:dphuman,densepose:parsingrcnn,densepose:amanet} or fully convolutional based frameworks \cite{pose_SunXLWang2019,pose_xiao2018simple}, since it is designed in an play-and-plug fashion. Next, we present the details of each module. 

\subsubsection{2D based Body Parsing Knowledge Representation}
To transfer the 2D based body parsing knowledge for densepose estimation, it is requisite to build a semantic representation for 2D visual nodes in $\mathit{G}$, which store prior 2D body-based parsing knowledge. To generate this kind of knowledge representation, a direct way is to construct independent feature representations. However, building such reference features often requires extra feature learning models (e.g., clustering or metric learning), leading to a huge computational burden. It is notable that parameters of 2D based body parsers characterize high-level parsing knowledge, since they can be used to estimate the location of human keypoints or parse the shape of body parts. Inspired by this, in this work the parsing knowledge is derived from three 2D based body parsers, including 1) the parameter $W_{loc}$ stored in bounding box classifier for human detection; 2) the parameter $W_{part}$ stored in pixel-wise classifier for body part segmentation; and 3) the parameter $W_{kpt}$ stored in the pixel-wise classifier for body keypoint estimation. For simplicity, we denote $W_{n} \in \mathbb{R}^{|C_n| \times D}: [W_{loc}, W_{part}, W_{kpt}]$ as the parameters copied from all 2D based body parsers, where $D$ is the size of each parsing parameter.

\subsubsection{Commonsense Relation Knowledge Graph} 
Parsing each 3D surface requires correlated 2D based body parsing knowledge. To achieve this, we construct the 2D-to-3D directed bipartite graph $\mathit{G}$ via two mechanisms: 1) Semantics Similarity $\mathcal{M}_{s} \in \mathbb{R}^{|C_s| \times |C_n|}$ and 2) Dependence Relationships $\mathcal{M}_{d} \in \mathbb{R}^{|C_s| \times |C_n|}$. 

We measure the semantics similarity between 2D visual nodes and 3D surface nodes via their linguistic distance, since previous work \cite{w2v:efficient} has demonstrated that word embeddings pre-trained on large scale language corpus involve strong semantics as well as linguistic relationships. In particular, we compute the cosine distance between the 2D visual nodes and the counterpart of 3D surface nodes. Formally, given the k-th 2D visual node and s-th surface node, their semantics similarity can be computed through Eq.\ref{equ.cos}: 
\begin{equation}
\begin{array}{lll}
\mathcal{M}_{s}(s,k) & = & \frac{\mathbf{W2V}(s)^{T}\mathbf{W2V}(k)}{\lVert\mathbf{W2V}(s)\rVert \lVert\mathbf{W2V}(k)\rVert} \\
\end{array}
\label{equ.cos} 
\end{equation}
where $\mathbf{W2V}(\cdot)$ is the Word2Vec function \cite{w2v:efficient} that extracts the word embedding from language words. 

As for the dependence relationship, we construct relation scores between 2D visual nodes and 3D surface nodes with the help of language corpus in the VG dataset \cite{visual_genome}. Specifically, we use all visual relationship annotations in VG, such as subject-verb-object relationship, and count frequent statistics for all edges $E$. Formally, the frequent statistics matrix is denoted as $\mathcal{M}_{f} \in R^{|C_s| \times |C_n|}$, and each element in $\mathcal{M}_{f}$ is rescaled into $(0, 1)$ by normalization. Next, we construct ``hard'' relations between 2D visual nodes and 3D surface nodes with the help of commonsense knowledge about human body structure. Specifically, we introduce an identification function $\mathcal{I}(s,k)$, where it outputs 1 $\mathit{iff}$ two nodes exist a compositional relation (e.g., head has eyes or body contains torso). Then, the dependence relationships $\mathcal{M}_{d} \in \mathbb{R}^{|C_s| \times |C_n|}$ is obtained through Eq.~\ref{equ.ind_rel}. 
\begin{equation}
\mathcal{M}_{d}(s,k)=\mathcal{M}_{f}(s,k) * \mathcal{I}(s,k)
\label{equ.ind_rel}
\end{equation}

Next, the 2D-to-3D directed bipartite graph $\mathit{G}$ is represented by a relational matrix $\mathcal{M}_{g} \in \mathbb{R}^{|C_n| \times |C_s|}$, which is derived from $\mathcal{M}_{s}$ and $\mathcal{M}_{d}$ via Eq.~\ref{equ.kg}:
\begin{equation}
\begin{array}{lll}
\mathcal{M}_{g} & = & \omega\mathcal{M}_{s} + \tau\mathcal{M}_{d} \\
\end{array}
\label{equ.kg} 
\end{equation}
where $\omega$ and $\tau$ are hyperparameters and set to 0.5. The associated parsing parameters $W^{a}_{n} = \mathcal{M}_{g}^{T}W_{n} \rightarrow \mathbb{R}^{|C_s| \times D}$ are calculated through a matrix multiplication operator.

\subsubsection{Knowledge Transformer} 
Given all associated parsing parameters $W^{a}_{n} \in \mathbb{R}^{|C_s| \times D}: [W^{a}_{loc},W^{a}_{part},W^{a}_{kpt}]$, we need to transform them to the parameters of the 3D body surface parser (i.e., pixel-wise surface classifier) for further densepose estimation. To achieve this, we design a small network with nonlinear mapping functions. Formally, we denote the parameters of the 3D body surface parser as $W_{s}$, and it can be generated through Eq.~\ref{equ.weigth_generator}:
\begin{equation}
\begin{array}{lll}
W_{s} & = &  \mathit{f}\left(\varPsi\left(W^{a}_{n}\right)W_{t}\right)W_{I} \\
\end{array}
\label{equ.weigth_generator} 
\end{equation}
where $\varPsi(\cdot)$ is the concatenation operation. The $W_{t} \in \mathbb{R}^{3D \times D}$ is the transform matrix for linear fusion and it can be adaptively updated during training. $\mathit{f}(\cdot)$ is the LeakyReLU \cite{activation_LReLU} nonlinear function with the negative slope of 0.2. The $W_{I}\in \mathbb{R}^{D \times D}$ is the linear transform matrix, which is used to generate final parameters of the 3D surface classifier. After jointly encoding parsing knowledge from the parameters of 2D based body parsers and structural priors from commonsense knowledge, the parameter $W_{s} \in \mathbb{R}^{|C_s| \times D}$ is utilized to parse 3D body surfaces accurately even it is trained under the condition of ``low-quality'' labels.

\subsubsection{DensePose Estimation}
Following previous works \cite{densepose:dphuman,densepose:parsingrcnn,densepose:amanet,densepose:ktn}, we apply 8 stacked convolutions to compute instance-level representations $\hat{f} \in R^{N \times D \times H_{r} \times W_{r}}$ with $D$ dimensions from $N$ propagated enhanced region features, and predict the following instance-level targets for densepose estimation:
\begin{enumerate}[(1)]
	\item Body mask of person body $P_{b}=\varphi_{b}(\hat{f}, W_{b})$, where $\varphi_{b}$ is the pixel-wise classifier with a parameter $W_{b}$.
	\item Body part mask $P_{bp}=\varphi_{bp}(\hat{f}, W_{part})$, where $\varphi_{bp}$ is the pixel-wise classifier with the parameter $W_{part}$.
	\item Heatmaps of body keypoints $P_{k}=\varphi_{k}(\hat{f}, W_{kpt})$, where $\varphi_{k}$ is the pixel-wise classifier with the parameter $W_{kpt}$.
	\item Segmentation maps of 3D body surface $P_{s} = \varphi_{s}(\hat{f}, W_{s})$, where $\varphi_{s}$ is the pixel-wise classifier with the parameter $W_{s}$ transferred from other parsing parameters (i.e., $W_{n}$).
	\item U coordinate maps of 3D body surface $P_{u} =\psi_{u}(\hat{f}, W_{u})$, where $\psi_{u}$ is the pixel-wise regressor with a parameter $W_{u}$.
	\item V coordinate maps of 3D body surface $P_{v} =\psi_{v}(\hat{f}, W_{v})$, where $\psi_{v}$ is the pixel-wise regressor with a parameter $W_{v}$.	
\end{enumerate}

\section{Experiments}
In this section, we first introduce the evaluation dataset and our implementation details in Sec.~\ref{sec.exp.setting}. Then, we perform ablation studies to investigate effects of each proposed component, involving an overall ablation study of two core components of KTN (Sec.~\ref{sec.exp.ktn}), specific ablation studies of proposed MID (Sec.~\ref{sec.exp.ktn_mid}) and detailed exploration on KTM (Sec.~\ref{sec.exp.ktnktm}). When performing specific ablation studies of core components, we particularly focus on the investigation of variants of proposed MID/KTM and the generalizability of core components, aiming to attain a further insight of proposed method. Finally, we compare the performance of our approach with the state-of-the-arts and provide an intuitive analysis with visualization results respectively in Sec.~\ref{sec.exp.comp_sota} and Sec.~\ref{sec.exp.vis}, which intuitively indicates the superiority as well as the limitation of the proposed method.

\subsection{Datasets and Implementation Details}
\label{sec.exp.setting}
\subsubsection{Dataset}
To assess effectiveness of proposed method in learning multi-person 2D-3D correspondences, we conduct experiments on DensePose-COCO dataset. Specifically, it is collected with dense correspondences manually annotated on the subset of COCO dataset \cite{dataset:coco}. For each human instance, it is annotated with 24 fine-grained body surfaces. The DensePose-COCO dataset contains about 50K humans annotations, each of which is annotated with 100 UV coordinates in average, and provides in total about 5 million manually annotated correspondences. Moreover, it is split into two subsets: training set and validation set with 32K images and 1.5k images, respectively.

\subsubsection{Implementation details}
We adopt SGD optimization algorithm to learn the network parameters, where the batch size is set to 8 and the momentum is 0.9. The initial learning rate is set as 0.01, and it respectively decreases to its 1/10 and 1/100 at the 200K-th and 240k-th iteration. Following previous work \cite{densepose:dphuman}, we report two types of evaluation metric over validation set: a) Average Precision (AP) and b) Average Recall (AR). Both of them are calculated at a number of geodesic point similarity (GPS) or object keypoints similarity (OKS), ranging from 0.5 to 0.95. Besides, other evaluation metrics (i.e., $AP_M$ and $AP_L$) for medium people and large people are also reported. To evaluate the performance of body segmentation, we follow \cite{densepose:parsingrcnn,humanparsing:understanding} to calculate standard mean intersection over union (mIoU), mean accuracy (mAcc) based on pixel-wise classification, Average Precision based on masks (i.e., AP$_{50}$ with IoU threshold equal to 0.5 and AP with IoU threshold ranging from 0.1 to 0.9, in increments of 0.1), and Percentage of Correctly parsed semantic Parts (PCP). We implement Knowledge Transfer Network (KTN) based on the Detectron2\footnote{https://github.com/facebookresearch/detectron2} on a server with 4 NVIDIA RTX GPUs, and adopt ResNet50 with FPN \cite{obj_det:FPN} as our backbone in the whole experiments unless otherwise mentioned. Other details are identical as in DensePose R-CNN \cite{densepose:dphuman}.

\subsection{Ablation Studies of Knowledge Transfer Network}
\label{sec.exp.ktn}
In this section, we focus on the investigation of proposed components of our knowledge transfer network (KTN). In particular, we choose the improved DensePose RCNN introduced in \cite{densepose:amanet} as our baseline and denote it as DensePose RCNN* for simplicity. Here, we compare 4 different settings:
\begin{enumerate}[(1)]
	\item DensePose RCNN*, which is the baseline model. 
	\item DensePose RCNN* with proposed multi-instances decoder (MID) but without the knowledge transfer machine (KTM); 
	\item DensePose RCNN* with the knowledge transfer machine (KTM) but without the multi-instances decoder (MID);
	\item DensePose RCNN* with all proposed components (i.e., MID and KTM), which is our knowledge transfer network (KTN).
\end{enumerate}
Besides, the MID/KTM proposed in our conference version \cite{densepose:ktn} is separately denoted as MIDv1 and KTMv1, and the upgraded version proposed in this work is respectively denoted as MIDv2 and KTMv2. Furthermore, the MIDv2 enhance feature representation power with feature rearranging operations that are not explored in the MIDv1. As for knowledge transfer machine (KTM), all correlated parameters of 2D parsers are transferred to 3D surface parser in KTMv2, while only keypoints parsing priors are utilized in KTMv1.
\begin{table}[ht]
	\caption{Ablation study. Investigating the effect of proposed modules in KTNet.}
	\label{Tab.abstudy_subcomp}
	\setlength{\belowcaptionskip}{0.cm}
	\setlength{\abovecaptionskip}{0.cm}
	\large
	\renewcommand\arraystretch{1.5}
	\resizebox{\linewidth}{!}{
		\begin{tabular}{|ccc|ccccc|}
			\hline
			\multicolumn{3}{|c|}{Setting}    & \multicolumn{5}{|c|}{Dense Pose Estimation} \\ \hline
			DensePose RCNN* & MIDv1 & KTMv1  & $AP$          & $AP_{50}$         & $AP_{75}$   &  $AP_{M}$ & $AP_{L}$    \\ \hline
			\checkmark	&        &     & 58.8 & 89.3 & 67.2 & 55.0  & 60.2        \\ \hline
			\checkmark	&   \checkmark        &    &     63.8        &      90.2        & 71.7 & 60.8 &    64.9          \\ \hline
			\checkmark	&        &   \checkmark     &         61.9          &     90.3        &  71.0 &  58.0 &   63.3    \\ \hline
			\checkmark	&    \checkmark  &   \checkmark     &    \textbf{66.5}         &    \textbf{91.5}          &   \textbf{75.5}    &  \textbf{61.9} &  \textbf{68.0}  \\
			\hline
			\hline
			DensePose RCNN* & MIDv2 & KTMv2  & $AP$          & $AP_{50}$         & $AP_{75}$   &  $AP_{M}$ & $AP_{L}$    \\ \hline
			\checkmark	&        &     & 58.8 & 89.3 & 67.2 & 55.0  & 60.2        \\ \hline
			\checkmark	&   \checkmark        &    &   64.4 & 90.8  & 73.6 &  60.2 &  65.7         \\ \hline
			\checkmark	&        &   \checkmark     &  63.4 & 91.6  & 72.2 &  61.0 &  64.8  \\ \hline
			\checkmark	&    \checkmark  &   \checkmark     &    \textbf{68.3}         &    \textbf{92.1}          &   \textbf{77.1}    &  \textbf{63.8} &  \textbf{70.0}  \\
			\hline
		\end{tabular}
	}
\end{table}

The experimental results are summarized in Tab.~\ref{Tab.abstudy_subcomp}. First, we observe that the performance of DensePose RCNN* is further improved when equipping with proposed components. In terms of the average precision, the DensePose RCNN* with MIDv1 outperforms the baseline by 5\%, which is improved further by 0.6\% when replacing MIDv1 with MIDv2. This indicates the effectiveness of rearranging operations (i.e. FA and IFA) for densepose estimation. With the KTMv1, the average precision is improved by 3.1\%, and it is further improved by 1.5\% with the help of KTMv2. In general, the baseline model with all proposed components obtains 66.5\% AP score and 68.3\% AP score, respectively. This demonstrates the overall effectiveness and importance of proposed components. Next, we perform extensive investigation of each component.

\subsection{Ablation Studies of Multi-instances Decoder}
\label{sec.exp.ktn_mid}
The multi-instances decoder consist of two main steps: instance completeness refining (ICR) and instance strengthening (IS), where the former is designed to generate unified feature representations, and the latter one is used to alleviate interference of background. In this section, we investigate effectiveness of each step by performing detailed component analysis in Sec.~\ref{sec.exp_mid_comp_details} and demonstrate the generalization capability of MID in Sec.~\ref{sec.exp_mid_generalize}. 
\subsubsection{Overall ablation studies of MID}
\label{sec.exp_mid_comp_details}
To investigate the effectiveness of each step, we compare 6 different settings based on modifications of Multi-instances decoder:
\begin{enumerate}[(1)]
	\item DensePose RCNN* without MID module.
	\item The multi-instances decoder with ICR step only but without IS step, where no feature rearranging operations are adopted in ICR step and it is designed for KTNetV1.
	\item The multi-instances decoder with ICR step only but without IS step, where feature rearranging operations are adopted in ICR step and it is designed for KTNetV2.
	\item The multi-instances decoder with IS step only but without ICR step.
	\item The multi-instances decoder with both ICR step and IS step, which is the MIDv1 module.
	\item The multi-instances decoder with both ICR step and IS step, which is the MIDv2 module.
\end{enumerate}
The experimental results are shown in Tab.~\ref{Tab.abstudy_mid}. Specifically, the mAP increases from 58.8\% to 63.0\% when applying MID with ICR step only but without feature rearranging operations, and it increases to 64.0\% when feature rearranging operations adopted in ICR step. Comparing row 1 with row 4, the mAP is increased by 3.4\%, which demonstrates the effectiveness of IS step. Furthermore, the MID with full proposed techniques achieves the best, implying that densepose estimation can benefit from ICR and IS, simultaneously.
\begin{table}[t]
	\caption{Ablation study based on modifications of MID. ICR denotes the abbreviation for Instance Completeness Refining, and IS is the abbreviation for the Instance Strengthening. FR denotes feature rearranging operations (i.e., feature adjustment (FA) and inverted feature adjustment (IFA)).}
	\label{Tab.abstudy_mid}
	\resizebox{\linewidth}{!}{
		\begin{tabular}{|cccc|ccc|}
			\hline
			Baseline & ICR w/o FR  & ICR w FR      & IS  &  $AP$ & $AP_{M}$ & $AP_{L}$    \\ \hline
			\checkmark &  &  &   & 58.8  & 55.0  & 60.2 \\ \hline
			\checkmark & \checkmark &  &   & 63.0  & 58.9  & 64.3 \\ \hline
			\checkmark &  & \checkmark  &   & 64.0  & 59.2  & 63.3 \\ \hline
			\checkmark &  &   & \checkmark  & 62.2  & 60.7  & 65.3 \\ \hline
			\checkmark & \checkmark &   & \checkmark  & 63.8  & \textbf{60.8} &   64.9 \\ \hline
			\checkmark &  & \checkmark  & \checkmark  & \textbf{64.4}  & 60.2 &   \textbf{65.7} \\ \hline
		\end{tabular}
	}
\end{table}

The MID generates three scale specific features by setting predefined dilation rates (i.e., 1,2,3) at ICR step, which captures instances with multi-scales. To investigate the effectiveness of such multi-scale setting, we perform a ablative study about the setting of dilation rate, as reported in Tab.~\ref{Tab.abstudy_mid_scale}. From the results, we can find that the performance of the densepose estimator on large instances increases, as the dilation rate increases. However, this cannot be hold on medium instances (See row 2,3). And the model with mixed dilation rates performs better than any models with only one dilation rate. Intuitively, large dilation rate has large visual receptive field that can perceive more complete visual context for large instance, but it may involve more irrelevant information especially when it extract features of instances with medium or small size. Therefore, it is necessary to equip the densepose estimator with multiple receptive fields (i.e., mixed dilation rates) for handling multi-scale instances.
\begin{table}[t]
	\caption{Ablation study based on modifications of scale setting.}
	\label{Tab.abstudy_mid_scale}
	\resizebox{\linewidth}{!}{
		\begin{tabular}{|c|ccccc|}
			\hline
			Dilation Rates & $AP$          & $AP_{50}$         & $AP_{75}$   &  $AP_{M}$ & $AP_{L}$    \\ \hline
			$\{1, 1, 1\}$ &  63.5  & 90.7  & 72.3 & 59.6  & 64.8 \\ \hline
			$\{2, 2, 2\}$ &  63.9  & 90.9  & 72.7 & 60.0  & 65.1 \\ \hline
			$\{3, 3, 3\}$ &  63.9  & \textbf{91.1}  & 72.4 & 59.3  & 65.2 \\ \hline
			$\{1, 2, 3\}$ &  \textbf{64.4}  & 90.8  & \textbf{73.6} & \textbf{60.2}  & \textbf{65.7} \\ \hline
		\end{tabular}
	}
\end{table}
\begin{table}[t]
	\caption{Generalization Study. Investigating the effect of MID.}
	\renewcommand\arraystretch{1.2}
	\resizebox{1.0\linewidth}{!}{
		\begin{tabular}{|cc|ccccc|}
			\hline
			\multicolumn{7}{|c|}{DensePose-COCO Body Segmentation} \\ 
			\hline
			DensePose RCNN* & MID  & $AP$   & $AP_{50}$ & $AP_{75}$ & $AP_{M}$ & $AP_{L}$  \\
			\hline
			\checkmark &            & 52.6  &  82.5   & 59.1  & 46.7 & 67.4    \\
			\checkmark & \checkmark & 53.7  &  82.5   & 60.1  & 47.5 & 68.9  \\
			\hline
			\multicolumn{2}{|c|}{Gains} & \multicolumn{1}{c}{\textbf{+1.1}} & \multicolumn{1}{c}{\textbf{+0.0}} & \multicolumn{1}{c}{\textbf{+1.0}} & \multicolumn{1}{c}{\textbf{+0.8}}  & \multicolumn{1}{c|}{\textbf{+1.5}} \\
			\hline
			\multicolumn{7}{|c|}{DensePose-COCO Part Segmentation} \\ 
			\hline
			DensePose RCNN* & MID  & $AP$  & $AP_{50}$ & mIoU & mAcc &  PCP$_{50}$  \\
			\hline
			\checkmark &            & 45.7 & 57.3  & 54.0 & 72.3  &    57.5  \\
			\checkmark & \checkmark & 52.6 & 70.6 & 60.0 & 79.3   &   73.5  \\
			\multicolumn{2}{|c|}{Gains} & \multicolumn{1}{c}{\textbf{+6.9}} & \multicolumn{1}{c}{\textbf{+13.3}} & \multicolumn{1}{c}{\textbf{+6.0}} & \multicolumn{1}{c}{\textbf{+7.0}}  &  \multicolumn{1}{c|}{\textbf{+16.0}} \\
			\hline
			\multicolumn{7}{|c|}{DensePose-COCO DensePose Estimation} \\ 
			\hline
			DensePose RCNN* & MID  & $AP$   & $AP_{50}$ & $AP_{75}$ & $AP_{M}$ & $AP_{L}$   \\
			\hline
			\checkmark &  & 58.8 & 89.3 & 67.2 & 55.0  & 60.2  \\
			\checkmark & \checkmark & 64.4  & 90.8  & 73.6 & 60.2  & 65.7   \\
			\hline
			\multicolumn{2}{|c|}{Gains} & \multicolumn{1}{c}{\textbf{+5.6}} & \multicolumn{1}{c}{\textbf{+1.5}} & \multicolumn{1}{c}{\textbf{+6.4}} & \multicolumn{1}{c}{\textbf{+5.2}}  & \multicolumn{1}{c|}{\textbf{+5.5}}
			\\
			\hline
			\multicolumn{7}{|c|}{DensePose-COCO Keypoint Detection} \\ 
			\hline
			DensePose RCNN* & MID  & $AP$   & $AP_{50}$ & $AP_{75}$ & $AP_{M}$ & $AP_{L}$   \\
			\hline
			\checkmark &            & 61.7 & 85.5  & 65.8 & 47.6  &  73.6  \\
			\checkmark & \checkmark & 63.5 & 87.0  & 68.2 & 53.4  &  72.8  \\
			\hline
			\multicolumn{2}{|c|}{Gains} & \multicolumn{1}{c}{\textbf{+1.8}} & \multicolumn{1}{c}{\textbf{+1.5}} & \multicolumn{1}{c}{\textbf{+2.4}} & \multicolumn{1}{c}{\textbf{+7.8}}  & \multicolumn{1}{c|}{\textbf{-0.8}} \\
			\hline
		\end{tabular}
	}
	\label{Tab:exp_mid_general}
\end{table}

\subsubsection{Generalizability of MID}
\label{sec.exp_mid_generalize}
The proposed Multi-instance decoder is not task-specific, and it can be applied in any human instance-level recognition tasks. In this section, we investigate the generalizability of the MID and apply it on several human instance-level tasks, including body segmentation, body part segmentation, densepose estimation and keypoints detection. In particular, we use the improved DensePose RCNN* \cite{densepose:amanet} as the baseline. The experimental results are summarized in Tab.~\ref{Tab:exp_mid_general}. From the results, we have following findings: 
1) The MID improves baseline model on all recognition tasks. 
2) The MID benefits Part-level visual tasks (i.e., densepose estimation and part segmentation) the most and improves them at least 5\% AP. 
3) The performance improvements on both body segmentation task and keypoints detection task are relative minor, which is 1.1\% and 1.8\%, respectively. 
Above findings demonstrate the generalizability of proposed MID, and suggest that enhancing the completeness of feature representation with background suppression can facilitate various human-centric recognition tasks, especially benefiting body parsing tasks, such as densepose estimation and body part segmentation, which requires more detailed information than other human-centric tasks. 

\begin{table}[t]
	\small
	\centering
	\caption{Ablation study. Investigating the role of 2D tasks.}
	\label{Tab.abstudy_extra_guidance}
	\renewcommand\arraystretch{1.3}
	\resizebox{1.0\linewidth}{!}{
		\begin{tabular}{|cccc|ccccc|}
			\hline
			\multicolumn{4}{|c|}{DensePose RCNN*}    & \multicolumn{5}{|c|}{Dense Pose Estimation} \\ \hline
			baseline    &  keypoints   &   BBox  & Parts & $AP$   & $AP_{50}$ & $AP_{75}$   & $AP_{M}$         & $AP_{L}$         \\ \hline
			\checkmark	&              &    &     & 58.8 & 89.3 & 67.2 & 55.0  & 60.2           \\ \hline
			\checkmark	&  \checkmark  &    &     & 61.9 & 90.3 & 71.0 &  58.0 & 63.3  \\    \hline
			\checkmark	&  \checkmark     & \checkmark    &  & 62.6  & 91.6 & 71.7 & \textbf{61.1} & 63.8         \\ \hline
			\checkmark	&  \checkmark     & \checkmark &  \checkmark   & \textbf{63.4} & \textbf{91.6}  & \textbf{72.2} &  61.0 &  \textbf{64.8}          \\ \hline
			\hline
			\multicolumn{4}{|c|}{DensePose RCNN* w MID}    & \multicolumn{5}{|c|}{Dense Pose Estimation} \\ \hline
			baseline    &  keypoints   &   BBox  & Parts &$AP$   & $AP_{50}$ & $AP_{75}$ & $AP_{M}$ & $AP_{L}$         \\ \hline
			\checkmark	&              &    &     & 64.4  & 90.8  & 73.6 & 60.2  & 65.7          \\ \hline
			\checkmark	&  \checkmark  &    &     & 67.3 & 91.6 & 76.8 & 62.7 & 68.8  \\    \hline
			\checkmark	&  \checkmark     & \checkmark    &  & 67.9      & 91.6        & 78.6             &  62.4  &  69.4         \\ \hline
			\checkmark	&  \checkmark     & \checkmark &  \checkmark   & \textbf{68.3}      & \textbf{92.1}   & \textbf{77.4}   &  \textbf{63.1}  &  \textbf{70.1}          \\ \hline
		\end{tabular}
	}
\end{table}

\subsection{Ablation Studies of Knowledge Transfer Machine} 
\label{sec.exp.ktnktm}
The proposed Knowledge Transfer Machine (KTM) is effective for densepose estimation and can be embedded in any existing densepose estimator without burden computation. In this section, we conduct ablative experiments to perform a detailed component analysis in Sec.~\ref{sec.exp_ktm_comp_details} and discuss various techniques for handling class-imbalance in Sec.~\ref{sec.exp_ktm_imbalance}. Moreover, we investigate the generalization capability of the proposed Knowledge Transfer Machine (KTM) in Sec.~\ref{sec.exp_ktm_generalize}.
\subsubsection{Overall ablation studies of KTM}
\label{sec.exp_ktm_comp_details}
In this section, we first quantitatively compare the effectiveness of transferring different 2D based tasks, namely, how much densepose estimator benefits from different 2D based tasks. Then, we investigate the sub-modules of KTM. 

To investigate the effects of 2D tasks, we set DensePose RCNN* and DensePose RCNN* with MIDv2 as our baselines.
The experimental results are shown in Tab.~\ref{Tab.abstudy_extra_guidance}. From Tab.\ref{Tab.abstudy_extra_guidance}, we can see that the average AP score can be increased by 4.6\% and 3.9\%, respectively for both baselines. Specifically, applying the KTM on keypoints detector brings major improvements (3.1\% and 2.9\%). This indicates the correlation between densepose and keypoints is stronger than other correlations, i.e., \{densepose, locations\} and \{densepose, body parts\}.

Next, we explore the effect of commonsense relation knowledge graph (CRKG) and knowledge transformer (KT). We also choose the DensePose RCNN* as our baseline and apply the KTM on keypoints detector only. Hence, we compare 4 different settings based on modifications of the KTM:
\begin{enumerate}[1)]
	\item DensePose RCNN* with both CRKG-S and KT, where the CRKG is implemented by Eq.~\ref{equ.weigth_generator} which aggregates relevant parsing parameters of 2D visual parsing tasks;
	\item DensePose RCNN* with both CRKG-A and KT, where the CRKG is defined as an averaging matrix that directly propagates all parsing parameters of 2D visual parsing tasks;
	\item DensePose RCNN* with CRKG-S only;
	\item DensePose RCNN* without KTM module. 
\end{enumerate}

From Tab.~\ref{Tab.abstudy_ktm}, we have the following observations: Firstly, the performance of densepose estimation decreases 4.7\% if we replace CRKG-S with a CRKG-A, and it is even worse than the baseline model without KTM. This implies the CRKG is the most essential component and directly passing parameters of keypoint detector is not proper. Second, the performance will drop to 61.1\% if KTM contains CRKG-S only, demonstrating the necessity of the knowledge transformer which further adapts the parameters of 2D body parsers to the 3D surface classifier. 

\begin{table}[t]
	\caption{Ablation study based on modifications of Knowledge Transfer Machine (KTM). CRKG-S denotes the proposed commonsense Relation Knowledge Graph, while CRKG-A is an averaging matrix. The KT is the knowledge transformer module. }
	\label{Tab.abstudy_ktm}
	\large
	\resizebox{\linewidth}{!}{
		\begin{tabular}{|c|ccccc|}
			\hline
			Modifications of KTM & $AP$          & $AP_{50}$         & $AP_{75}$   &  $AP_{M}$ & $AP_{L}$    \\ \hline
			CRKG-S and KT &       \textbf{61.9}          &     \textbf{90.3}        &  \textbf{71.0} &  \textbf{58.0} &   \textbf{63.3}      \\ \hline
			CRKG-A and KT     & 57.2 & 89.7 & 65.8 & 53.3  & 58.6       \\ \hline
			CRKG-S only  &     61.1        &  90.5          &  69.2   &  58.1 & 62.6      \\ \hline
			No KTM  & 58.8 & 89.3 & 67.2 & 55.0  & 60.2       \\ \hline
			
		\end{tabular}
	}
\end{table}

\begin{table}[ht]
	\caption{Comparison with various techniques of alleviating class-imbalance on densepose estimation task.}
	\label{Tab.abstudy_imbalance}
	\large
	\renewcommand\arraystretch{1.2}
	\resizebox{\linewidth}{!}{
		\begin{tabular}{|c|ccccc|}
			\hline
			Methods            &     $AP$      &   $AP_{50}$   &   $AP_{75}$   &   $AP_{M}$    &  $AP_{L}$    \\ \hline
			KTN w/o KTM        &     63.8      &    90.2       &     71.7      &   60.8        &   64.9       \\ \hline
			w re-weighting       &     64.2      &    91.3       &     72.6      &   60.0        &   65.3       \\ \hline
			w re-sampling        &     64.8      &    91.7       &     72.5      &   60.1        &   65.9       \\ \hline
			w OHEM triplet loss  &     64.1      &    91.4       &     72.1      &   60.2        & 65.1         \\ \hline
			w KTM                & \textbf{66.5} & \textbf{91.5} & \textbf{75.5} & \textbf{61.9} & \textbf{68.0}\\ \hline
		\end{tabular}
	}
\end{table}
\subsubsection{Ablation studies of KTM for class-imbalance issue}
\label{sec.exp_ktm_imbalance}
In this section, we compare proposed KTM with three mature techniques for handling class-imbalance problem, including:
\begin{enumerate}[(1)]
	\item Re-weighting method, where the training weight of each surface category is decided by the proportion of its samples. 
	\item Re-sampling method, where samples of minor surface categories are picked with higher possibility than those of major categories. 
	\item OHEM triplet loss, where we apply an extra triplet loss on minor surface categories during the training, which is a special case of on-line hard example mining.
\end{enumerate}

\begin{figure}[ht]
	\centering
	\begin{center}
		\includegraphics[width=1.0\linewidth]{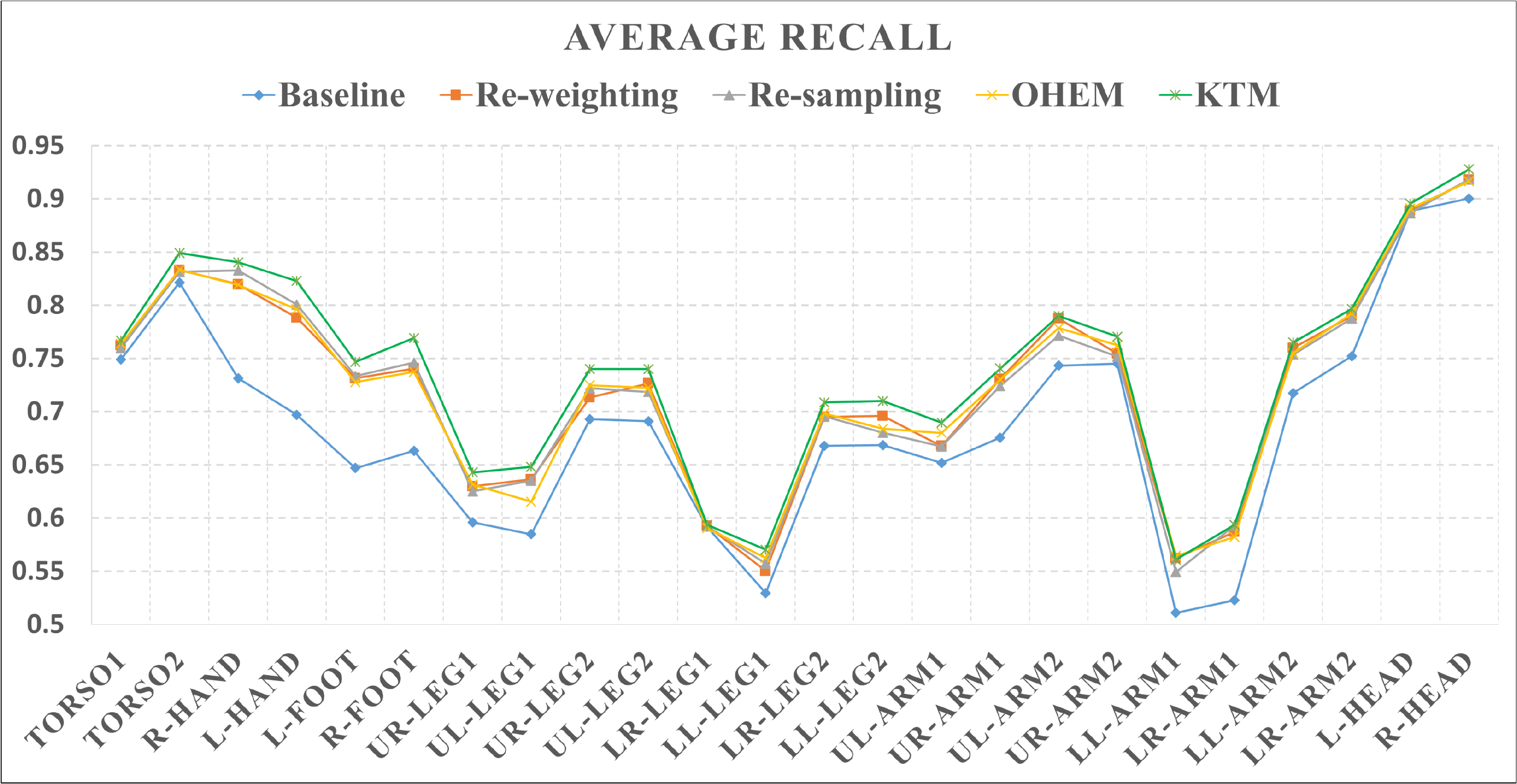}
	\end{center}
	\caption{Category based average recalls obtained by various methods for handling the imbalance problem, including 1) re-weighting; 2) re-sampling; 3) on-line hard example mining (OHEM) and 4) proposed knowledge transfer machine (KTM). Zoom in for a better view.}
	\label{fig:imbalance_methods}
\end{figure}

For a fair comparison, we use our knowledge transfer network without KTM as the baseline model. The experimental results are summarized in Tab.~\ref{Tab.abstudy_imbalance}, which illustrates that the performance can be improved when applying any technique for alleviating class-imbalance. In terms of the average precision, the improvements obtained from mature techniques are 0.3\%$\sim$1.0\%, while the AP gains are 2.7\% when using KTM. Fig.~\ref{fig:imbalance_methods} shows category based average recalls obtained by the aforementioned techniques. From Fig.~\ref{fig:imbalance_methods}, we can see that major improvements when applying mature techniques only come from major classes such as Torso, hand and arm. However, the average recall of most classes, including minor ones such as leg and foot, consistently benefit from the KTM technique. It demonstrates the effectiveness of the KTM for handling the class-imbalance problem that exists in densepose estimation.

\subsubsection{Generalizability of KTM}
\label{sec.exp_ktm_generalize}
We investigate the generalization capability of proposed Knowledge Transfer Machine by adding it to two types of densepose estimation system: 1) RCNN based approaches and 2) Fully-convolutional frameworks. In particular, we choose three different RCNN based models that achieve state-of-the-art performance for densepose estimation, including DensePose R-CNN \cite{densepose:dphuman}, Parsing-RCNN \cite{densepose:parsingrcnn} and AMANet \cite{densepose:amanet}. As for Fully-convolution based approaches, we use two different pose estimators as our baselines that are SimpleNet \cite{pose_xiao2018simple} and HRNet \cite{pose_SunXLWang2019} respectively.

The experimental results are shown in Tab.~\ref{tab:abstudy_generalize}. As can be seen, DensePose R-CNN with proposed KTM boosts the AP scores from 51.9\% to 55.2\%, while Parsing-RCNN with KTM has 3.9\% improvements in terms of mAP metric. Furthermore, the performance of AMA-net can be improved by up to 2.0\% with the help of KTM. A similar performance improvement can also be observed for Fully-convolution based approaches, which is 2.8\% and 1.0\% for SimpleNet and HRNet, respectively. This clearly indicates the effectiveness and generalizability of the proposed KTM.

\begin{table*}[t]
	\centering
	\caption{Ablation study. Investigating the generalizability of Knowledge Transfer Machine.}
	\renewcommand\arraystretch{1.2}
	\begin{tabular}{|l|c|c|c|c|c|c|c|c|c|c|}
		
		\hline
		Method                                                 & AP   & AP$_{50}$ & AP$_{75}$ & AP$_{M}$ & AP$_{L}$ & AR   & AR$_{50}$ & AR$_{75}$ & AR$_{M}$ & AR$_{L}$ \\
		\hline
		\multicolumn{11}{|c|}{RCNN-based methods}  \\
		\hline
		DensePose R-CNN \cite{densepose:dphuman}                                           & 51.9 & 85.5 & 54.7 & 39.4  & 53.9  & 61.1 & 89.7 & 65.5 & 42.0  & 62.4  \\
		+ KTM                                      & $55.2^{+3.3}$ & $88.7^{+3.2}$ & $61.9^{+7.2}$ & $53.4^{+14.0}$  & $56.5^{+2.6}$  & $63.8^{+2.7}$ & $92.7^{+3.0}$ & $71.3^{+5.8}$ & $54.8^{+12.8}$  & $64.4^{+2.0}$  \\ \hline
		Parsing R-CNN \cite{densepose:parsingrcnn}                                           &  58.3 & 90.1 & 66.9 & 51.8  & 61.9  & - & - & - & -  & -  \\
		+ KTM                                       & $62.2^{+3.9}$ & $90.7^{+0.6}$ & $70.2^{+3.3}$ & $57.9^{+6.1}$   & $63.6^{+1.7}$  & $70.4$ & $94.3$ & $77.8$ & $59.2$  & $71.1$  \\ \hline
		AMA-net \cite{densepose:amanet}                                      & 64.1 & 91.4 & 72.9 & 59.3  & 65.3  & 71.6 & 94.7 & 79.8 & 61.3  & 72.3  \\
		+ KTM    & $66.1^{+2.0}$ & $91.8^{+0.4}$ & $75.2^{+2.3}$ & $62.9^{+3.6}$  & $67.5^{+2.2}$  & $74.2^{+2.6}$    & $95.3^{+0.6}$    & $82.6^{+2.8}$    & $65.3^{+4.0}$     & $74.8^{+2.5}$     \\ \hline
		\hline
		\multicolumn{11}{|c|}{Fully-convolutional methods}  \\
		\hline
		Simple \cite{pose_xiao2018simple}                                    & 60.1 & 90.2 & 67.2 & 56.4  & 61.5  & 68.4 & 94.2 & 75.9 & 57.8  & 69.0  \\
		+ KTM    & $62.9^{+2.8}$ & $92.5^{+2.3}$ & $73.6^{+6.4}$ & $60.7^{+4.3}$  & $63.8^{+2.3}$  & $70.2^{+1.8}$    & $95.8^{+1.6}$    & $80.5^{+4.6}$    & $62.6^{+4.8}$     & $70.7^{+1.7}$     \\ \hline
		HRNet \cite{pose_SunXLWang2019}                                      & 65.1 & 92.9 & 76.8 & 62.4  & 66.2  & 72.3 & 96.1 & 83.4 & 64.5  & 72.8  \\
		+ KTM    & $66.1^{+1.0}$ & $92.6^{-0.3}$ & $78.8^{+2.0}$ & $64.3^{+1.9}$  & $67.2^{+1.0}$  & $73.4^{+1.1}$    & $96.1^{+0.0}$    & $85.0^{+1.6}$    & $66.5^{+2.0}$     & $73.8^{+1.0}$     \\ \hline
		
	\end{tabular}
	\label{tab:abstudy_generalize}
\end{table*}

\begin{table*}[t]
	\centering
	\caption{Exploration study of fully-convolutional densepose estimation model.}
	\label{Tab.abstudy_fullyconv}
	\renewcommand\arraystretch{1.2}
	\begin{tabular}{|ccccc|ccccc|}
		\hline
		\multicolumn{5}{|c|}{Setting}    & \multicolumn{5}{|c|}{Dense Pose Estimation} \\ \hline
		Method & Initialization & MID & Data Aug  & KTM & $AP$          & $AP_{50}$         & $AP_{75}$   &  $AP_{M}$ & $AP_{L}$    \\ \hline
		Simple\cite{pose_xiao2018simple}	&   ImageNet     &    &    &    & 53.8   &   90.9  &  59.5   & 55.7 & 54.5       \\ \hline
		Simple\cite{pose_xiao2018simple}	&   COCO     &    &    &     &    56.6         &    91.9          & 67.1 & 56.5 & 57.4             \\ \hline
		Simple\cite{pose_xiao2018simple}	&   COCO     &  \checkmark &   &    &   58.8          &   92.1         &   68.6  & 57.6 & 59.8      \\ \hline
		Simple\cite{pose_xiao2018simple}	&  COCO   & \checkmark  &  \checkmark   &    &    60.1         &   90.2 & 67.2 & 56.4  & 61.5   \\\hline
		Simple\cite{pose_xiao2018simple}	&   COCO    & \checkmark   & \checkmark   & \checkmark    &     62.9        &   92.5          & 73.6  & 60.7 & 63.8        \\ \hline
		HRNet\cite{pose_SunXLWang2019}	&   COCO     &  \checkmark &  \checkmark   &    & 65.1 & \textbf{92.9} & 76.8 & 62.4  & 66.2    \\ \hline
		HRNet\cite{pose_SunXLWang2019}	&    COCO   &  \checkmark  &  \checkmark  &   \checkmark &   \textbf{66.1}         &     92.6       &  \textbf{78.8}  &  \textbf{64.3} & \textbf{67.2}    \\ \hline
	\end{tabular}
\end{table*}
\begin{table*}[t]
	\centering
	\caption{Comparison with the state-of-the-art methods on dense pose estimation task.}
	\renewcommand\arraystretch{1.2}
	\begin{tabular}{|c|c|c|c|c|c|c|c|c|c|c|c|}
		
		\hline
		\multicolumn{1}{|c|}{Method}    &   Backbone   & AP   & AP$_{50}$ & AP$_{75}$ & AP$_{M}$ & AP$_{L}$ & AR   & AR$_{50}$ & AR$_{75}$ & AR$_{M}$ & AR$_{L}$ \\
		\hline
		DensePose R-CNN \cite{densepose:dphuman}                & ResNet50          & 51.0 & 83.5 & 54.2 & 39.4  & 53.1  & 60.1 & 88.5 & 64.5 & 42.0  & 61.3  \\
		DensePose R-CNN \cite{densepose:dphuman}                & ResNet101         & 51.8 & 83.7 & 56.3 & 42.2  & 53.8  & 61.1 & 88.9 & 66.4 & 45.3  & 62.1  \\
		DensePose R-CNN+keypoints \cite{densepose:dphuman}      & ResNet50          & 52.8 & 85.6 & 56.2 & 42.2  & 54.7  & 62.6 & 89.8 & 67.7 & 45.4  & 63.7  \\
		DensePose R-CNN+masks \cite{densepose:dphuman}          & ResNet50          & 51.9 & 85.5 & 54.7 & 39.4  & 53.9  & 61.1 & 89.7 & 65.5 & 42.0  & 62.4  \\
		DensePose R-CNN+cascade \cite{densepose:dphuman}        & ResNet50          & 51.6 & 83.9 & 55.2 & 41.9  & 53.4  & 60.4 & 88.9 & 67.7 & 45.4  & 63.7  \\
		DensePose R-CNN+cascade+masks \cite{densepose:dphuman}  & ResNet50          & 52.8 & 85.5 & 56.1 & 40.3  & 54.6  & 62.0 & 89.7 & 67.0 & 42.4  & 63.3  \\
		DensePose R-CNN+cascade+keypoints\cite{densepose:dphuman} & ResNet50        & 55.8 & 87.5 & 61.2 & 48.4  & 57.1  & 63.9 & 91.0 & 69.7 & 50.3  & 64.8  \\\hline
		Slim DP \cite{densepose:slimdp}                         & ResNeXt101        & 55.5 & 89.1 & 60.8 & 50.7  & 56.8  & 63.2 & 92.6 & 69.6 & 51.8  & 64.0 \\ 
		Slim DP \cite{densepose:slimdp}                         & Hourglass         & 57.3 & 88.4 & 63.9 & 57.6  & 58.2  & 65.8 & 92.6 & 73.0 & 59.6  & 66.2 \\ \hline
		SimPose \cite{densepose:simpose}                        & ResNet101         & 57.3 & 88.4 & 67.3 & 60.1  & 59.3  & 66.4 & 95.1 & 77.8 & 62.4  & 66.7 \\ \hline
		Parsing R-CNN    \cite{densepose:parsingrcnn}           & ResNet50          & 55.0 & 87.6 & 59.8 & 50.6  & 56.6  & -    & -    & -    & -     & -     \\
		Parsing R-CNN+keypoints    \cite{densepose:parsingrcnn} & ResNet50          & 58.3 & 90.1 & 66.9 & 51.8  & 61.9  & -    & -    & -    & -     & -     \\ 
		Parsing R-CNN    \cite{densepose:parsingrcnn}           & ResNeXt101        & 59.1 & 91.0 & 66.9 & 51.8  & 61.9  & -    & -    & -    & -     & -     \\
		Parsing R-CNN+keypoints    \cite{densepose:parsingrcnn} & ResNeXt101        & 61.6 & 91.6 & 72.3 & 54.8  & 64.8  & -    & -    & -    & -     & -     \\  \hline
		AMA-net \cite{densepose:amanet}                         & ResNet50          & 64.1 & 91.4 & 72.9 & 59.3  & 65.3  & 71.6 & 94.7 & 79.8 & 61.3  & 72.3 \\ \hline  
		DensePose R-CNN+uncertainty\cite{densepose:uncertainty} & ResNet50          & 64.2 & 91.2 & 72.6 & 58.7  & 65.4  & 71.7 & 94.6 & 79.2 & 60.3  & 72.4 \\ 
		DensePose R-CNN+uncertainty\cite{densepose:uncertainty} & ResNet101         & 64.9 & 91.9 & 72.8 & 60.6  & 66.0  & 72.5 & 94.9 & 79.7 & 62.5  & 73.2 \\ \hline
		KTNetV1 \cite{densepose:ktn}                            & ResNet50          & 66.5 & 91.5 & 75.5 & 61.9  & 68.0 & 74.2 & 95.2 & 82.3 & 64.2  & 74.9 \\ 
		KTNetV1 \cite{densepose:ktn}                            & ResNeXt101        & 67.5 & 92.0 & 75.4 & 63.8  & 68.8 & 74.9 & 95.3 & 81.9 & 66.0  & 75.5 \\
		KTNetV1 \cite{densepose:ktn}                            & HRNetW32          & 69.6 & 92.8 & 80.0 & 65.4  & 70.8 & 76.7 & 95.5 & 85.1 & 68.2  & 77.3 \\
		KTNetV1 \cite{densepose:ktn}                            & HRNetW48          & 69.6 & 92.9 & 78.6 & 68.1  & 70.7 & 77.0 & 95.6 & 84.8 & 70.1  & 77.5 \\ \hline
		KTNetV2                                                 & ResNet50          & 68.3 & 92.1 & 77.1 & 63.8  & 70.0 & 76.3 & 96.2 & 84.0 & 67.9  & 76.8 \\
		KTNetV2                                                 & ResNeXt101        & 69.2 & 92.5 & 78.8 & 67.7  & 70.5 & 76.7 & 96.2 & 84.4 & 69.9  & 77.2 \\
		KTNetV2                                                 & HRNetW32          & 71.8 & \textbf{93.7} & 81.5 & 67.3  & \textbf{73.1} & 78.5 & \textbf{96.7} & 86.6 & 69.5  & 79.1 \\ 
		KTNetV2                                                 & HRNetW48          & \textbf{71.9} & 93.3 & \textbf{82.7} & \textbf{69.3}  & 72.9 & \textbf{78.6} & 96.3 & \textbf{87.1} & \textbf{71.1}  & \textbf{79.2} \\		
		\hline
	\end{tabular}
	\label{tab:com_stoa}
\end{table*}
\begin{figure*}[ht]
	\setlength{\belowcaptionskip}{-0.3cm}%
	\centering
	\includegraphics[width=0.95\linewidth]{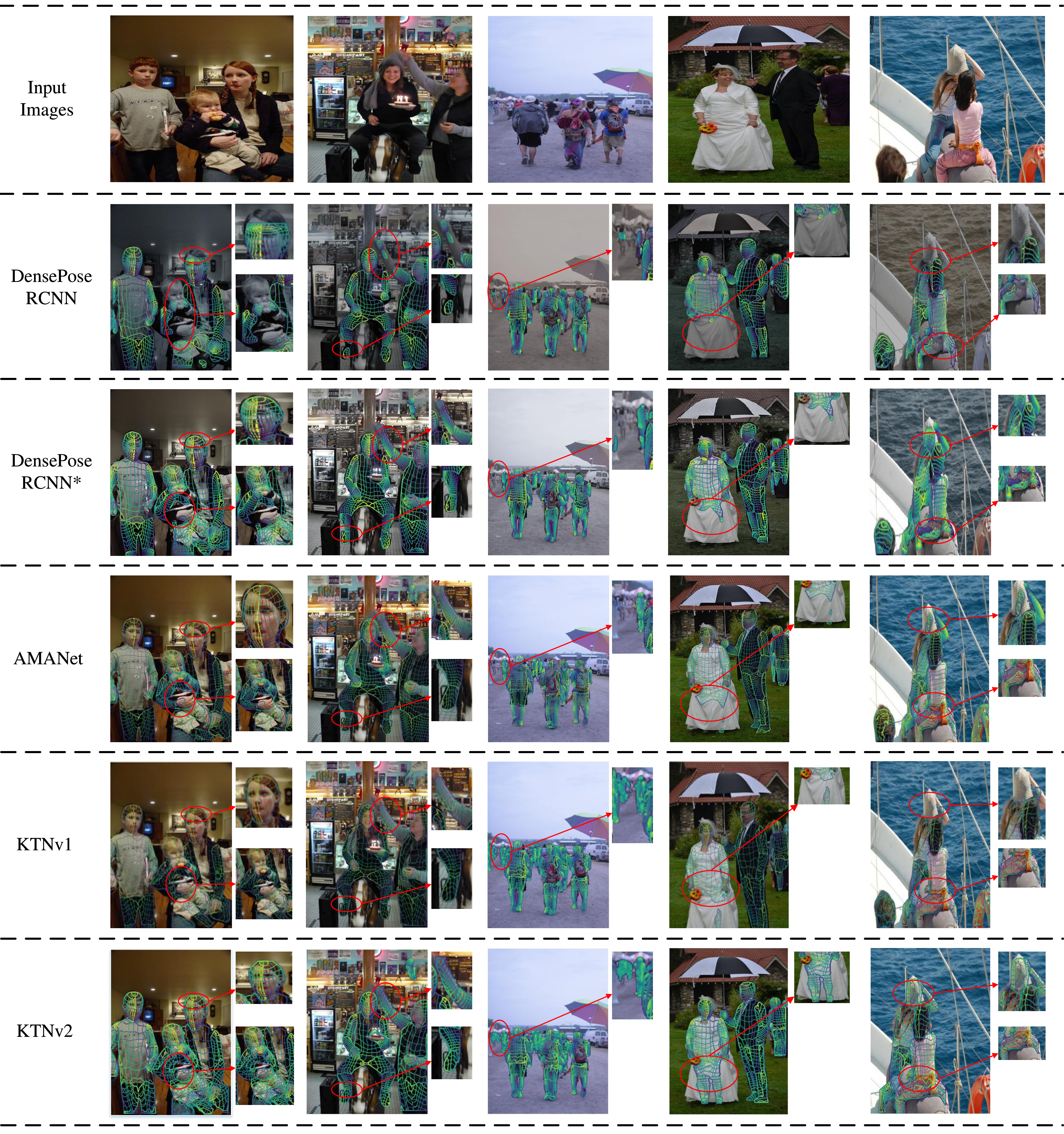}
	\centering
	\caption{Qualitative comparison. From the top to bottom: the original input images, UV coordinates from DensePose RCNN\cite{densepose:dphuman}, DensePose-RCNN*\cite{densepose:amanet}, AMANet\cite{densepose:amanet}, KTNetV1\cite{densepose:ktn} and KTNetV2, respectively. Red circles spot the difference among the predictions.}
	\label{fig.pred_comp}
\end{figure*}

\begin{table*}[ht]
	\centering
	
	\caption{Category based evaluation results, involving: (1) average recall; (2) mean square error and (3) UV geodesic distance.}
	\renewcommand\arraystretch{1.2}
	\resizebox{1.\linewidth}{!}{
		\begin{tabular}{|c|c|cccccccccc|}
			\hline
			Method &  AR  & Head  & Torso & R-Arm & L-Arm & R-Hand & L-Hand & R-Leg & L-Leg & R-Foot & L-Foot  \\ \hline
			Baseline        & 68.5 & 89.4  & 78.5  & 67.4  & 65.6  & 73.1   & 69.7   & 63.7  & 61.8  & 66.3   & 64.7 \\
			KTN (ResNet50)  & 73.7 & 91.2  & 80.8  & 72.5  & 70.1  & 84.0   & 82.3   & 67.1  & 66.7  & 76.9   & 74.7 \\
			KTN (HRNetW32)  & 76.2 & 91.6  & 81.9  & 75.0  & 72.1  & 87.2   & 85.0   & 70.5  & 68.9  & 81.8   & 80.8 \\
			KTN (HRNetW48)  & 75.0 & 90.8  & 81.2  & 73.6  & 71.4  & 86.1   & 82.5   & 69.5  & 68.1  & 79.2   & 78.1 \\ \hline
			\hline
			Method & U \textit{MSE}  & Head   & Torso  & R-Arm  & L-Arm  & R-Hand  & L-Hand  & R-Leg  & L-Leg  & R-Foot & L-Foot  \\ \hline
			Baseline        & 0.079           & 0.066  & 0.047  & 0.077  & 0.083  & 0.110   & 0.104   & 0.083  & 0.084  & 0.075  & 0.068 \\
			KTN (ResNet50)  & 0.079           & 0.066  & 0.050  & 0.076  & 0.083  & 0.117   & 0.113   & 0.082  & 0.083  & 0.076  & 0.067 \\
			KTN (HRNetW32)  & 0.075           & 0.063  & 0.045  & 0.072  & 0.079  & 0.106   & 0.099   & 0.081  & 0.080  & 0.072  & 0.063 \\
			KTN (HRNetW48)  & 0.075           & 0.064  & 0.045  & 0.073  & 0.080  & 0.108   & 0.103   & 0.079  & 0.080  & 0.072  & 0.063 \\ \hline
			\hline
			Method & V \textit{MSE}  & Head   & Torso  & R-Arm  & L-Arm  & R-Hand  & L-Hand  & R-Leg  & L-Leg  & R-Foot & L-Foot  \\ \hline
			Baseline        & 0.085           & 0.045  & 0.063  & 0.091  & 0.100  & 0.093   & 0.095   & 0.088  & 0.087  & 0.080  & 0.076 \\
			KTN (ResNet50)  & 0.083           & 0.042  & 0.066  & 0.087  & 0.099  & 0.096   & 0.099   & 0.085  & 0.085  & 0.082  & 0.077 \\
			KTN (HRNetW32)  & 0.080           & 0.041  & 0.045  & 0.085  & 0.096  & 0.083   & 0.086   & 0.083  & 0.082  & 0.077  & 0.072 \\
			KTN (HRNetW48)  & 0.080           & 0.041  & 0.063  & 0.085  & 0.096  & 0.085   & 0.086   & 0.083  & 0.083  & 0.077  & 0.073 \\ \hline
			\hline
			Method & UV \textit{GD}  & Head   & Torso  & R-Arm  & L-Arm  & R-Hand  & L-Hand  & R-Leg  & L-Leg  & R-Foot & L-Foot  \\ \hline
			Baseline        & 0.085           & 0.047  & 0.107  & 0.061  & 0.063  & 0.062   & 0.067   & 0.108  & 0.109  & 0.068  & 0.077 \\
			KTN (ResNet50)  & 0.083           & 0.040  & 0.095  & 0.051  & 0.055  & 0.056   & 0.059   & 0.096  & 0.096  & 0.057  & 0.061 \\
			KTN (HRNetW32)  & 0.080           & 0.039  & 0.090  & 0.049  & 0.053  & 0.050   & 0.053   & 0.091  & 0.093  & 0.052  & 0.052 \\
			KTN (HRNetW48)  & 0.080           & 0.040  & 0.095  & 0.051  & 0.054  & 0.052   & 0.056   & 0.093  & 0.095  & 0.054  & 0.056 \\ \hline
		\end{tabular}
		\label{tab.predict_analysis}
	}
\end{table*}

\subsection{Fully-convolutional Densepose Estimator} 
Fully-convolution based approaches are driven by two-stage pipeline and achieve superior results especially for human keypoints estimation. They mainly differ from RCNN based methods in terms of their region-based methods. For example, RCNN-based approaches \cite{ins_seg:mask_rcnn,densepose:dphuman} provide a unified solution that unifies the object detector and the single-person parser into a complete model, while fully-convolution based approaches \cite{pose_SunXLWang2019,pose_xiao2018simple} use two independent fully-convolutional networks for object detection and single-person parser respectively. Therefore, fully-convolution based methods often have better performance but require more computation resources, when comparing with RCNN based approaches. In this section, we provide a solid exploration study of fully-convolutional densepose estimator with our proposed components. More specifically, we investigate not only the effect of proposed components but also the training strategies, such as initialization methods and data augmentation, which are crucial for training fully-convolution based models from insufficient labels. Here, we use the model weights pre-trained either from ImageNet or COCO for initialization. As for data augmentation, we simply expand correspondence labels by linear interpolation as introduced in \cite{densepose:dphuman}. Moreover, we simply re-implement the methods of SimpleNet and HRNet by modifying their output for densepose estimation, and set them as the baselines of fully-convolutional densepose estimation models.  

The experimental results are summarized in Tab.~\ref{Tab.abstudy_fullyconv}. First, we see that the performance of densepose estimator initialized from ImageNet model is worse than the one initialized from COCO model, which implies carefully designed initialization is necessary. The performance can be improved when suppressing the interference of backgrounds by adding proposed MID module, which obtains 58.8\% AP scores. Moreover, the performance is boosted to 62.9\% with 4.1\% gains when applying the data augmentation strategy and proposed Knowledge Transfer Machine, demonstrating the importance of learning strategy. Similar gains can be observed when using HRNet as the backbone (See row 6 and row 7 in Tab.~\ref{Tab.abstudy_fullyconv}), showing the generalizability of our proposed method. 

\subsection{Comparison with State-of-the-art Methods}
\label{sec.exp.comp_sota}
After exploring the effectiveness of knowledge transfer network and the generalization capability of proposed components (MID and KTM), we compare our proposed KTN with state-of-the-art methods \cite{densepose:dphuman,densepose:parsingrcnn,densepose:amanet,densepose:slimdp,densepose:simpose,densepose:uncertainty} and their variants with the additional guidances, such as COCO keypoints. The comparison results are summarized in Tab.~\ref{tab:com_stoa}. First, we can observe that the performance of both DensePose R-CNN and Parsing R-CNN is significantly boosted when incorporating additional guidance of keypoints. Specifically, the performance of DensePose R-CNN can be increased by 4\% in terms of AP score, while the counterpart of Parsing R-CNN can be 2.5\%. Moreover, our KTNet with the help of proposed components achieves the best performance in all the measure metrics among all the comparison methods, reaching 68.3\% AP and 76.3\% AR, which is 1.8\% AP, 2.1\% AR higher than our conference version \cite{densepose:ktn}. In addition, it can achieve higher performance on all metrics after using large models such as ResNeXt101 and HRNet. Specifically, it achieves 69.2\% AP, 71.8\% AP and 71.9\% AP when applying ResNeXt101, HRNetW32 and HRNetW48, respectively. The consistent improvement indicates the importance of enhancing the completeness of feature representation when estimating human densepose. This is because large models such as ResNeXt101 or HRNet have a large visual perceptive field and keep features at high resolution with negligible spatial loss.

\subsection{Visualization} 
\label{sec.exp.vis}
In this section, we provide an intuitive analysis by visualizing the prediction results obtained from proposed KTN.

\noindent\textbf{Qualitative Comparison:} Fig.~\ref{fig.pred_comp} reports the generated 3D surface contours by DensePose RCNN\cite{densepose:dphuman}, DensePose RCNN* which is our baseline model, AMANet \cite{densepose:amanet}, KTNetV1 \cite{densepose:ktn} and KTNetV2. The first row shows input images. The second, third and fourth rows respectively show the 3D surface contours produced by DensePose-RCNN, DensePose-RCNN* and AMANet. The fifth and sixth row shows the results from KTNetV1 and KTNetV2. 
From the visualization comparison, we have the following observations:
1) Most approaches produce incomplete densepose estimations under the complex scenarios, such as heavily intertwined occlusions (columns 1, 5), diverse postures (column 2), small instances (column 3) and self-occlusions (column 4).
2) Densepose estimations obtained from KTN based methods are well-shaped for each instance, but the estimated contours are distorted under the condition of severe occlusion.
The visualization comparisons quantitatively reveal the effectiveness of the proposed KTN for producing complete densepose estimations. Meanwhile, the deficiency at UV regression is also exposed when it comes to ``heavy occlusion''.

\noindent\textbf{Analysis of Errors:} The densepose estimation involves two sub-tasks: surface classification and surface-specific regression. The improvements brought by the proposed method may come from either one of them or the both. To figure out how much has been improved for the two subtasks, we perform the point-wise evaluation on the proposed KTN across various backbones. The quantitative results and corresponding visualizations are summarized in Tab.~\ref{tab.predict_analysis}, respectively. From the results, we observe that: 
1) In terms of surface classification, the KTN based methods perform better when compared with baseline model. 
2) For UV regression task, the improvement come from proposed method is minor. It implies that the proposed KTN facilitates surface classification task but provides little help to UV based regression task.

\section{Conclusion}
In this paper, we propose the knowledge transfer network for dense pose estimation. To handle the issue of incomplete estimations, we propose multi-instances decoder (MID) to maintain feature resolutions and suppress the effect of backgrounds. Then the knowledge transfer machine (KTM) is introduced to address the issue of learning densepose estimation from incomplete labels. Extensive experiments on DensePose-COCO benchmark suggest that our approach consistently outperforms state-of-the-art competitors, indicating the effectiveness and generalizability of our approach.

\bibliographystyle{IEEEtran}
\bibliography{bib}

\section*{Appendix A: The effect of 2D based body parser}
The extended version of proposed Knowledge Transfer Network (KTN) depends on three 2D based body parsers. In this section, we gauge the contribution of these parsers, that is, which 2D based body parsing knowledge is the main contribution to the densepose estimation. Following the experimental settings as stated in regular manuscript, we adopt two models for this investigation, i.e., DensePose RCNN$^*$ and DensePose RCNN$^*$ with MID. The experimental results are reported in Tab.~\ref{Tab.supl_abstudy_extra_guidance}. From the results, we observe that location parser brings relative minor contribution (58.8\% vs 60.0\% and 64.4\% vs 66.0\%) to the densepose estimation, while part parser as well as keypoint parser improve densepose parser by a large margin, i.e., from 58.8\% to 62.2\%/61.9\% and from 64.4\% to 67.6\%/67.3\%. We conjecture the possible reason behind this is that the part segmentation or keypoint estimation shares a common characteristic with the densepose estimation, that is, the essential problem of these tasks is the part-level dense classification problem.

\begin{table}[hb]
	\small
	\centering
	\caption{Ablation study. Investigating the role of 2D tasks.}
	
	\renewcommand\arraystretch{1.3}
	\resizebox{1.0\linewidth}{!}{
		\begin{tabular}{|cccc|ccccc|}
			\hline
			\multicolumn{4}{|c|}{DensePose RCNN*}    & \multicolumn{5}{|c|}{Dense Pose Estimation} \\ \hline
			baseline    &   BBox  & Parts &  keypoints &$AP$   & $AP_{50}$ & $AP_{75}$ & $AP_{M}$ & $AP_{L}$         \\ \hline
			\checkmark	&              &    &     & 58.8 & 89.3 & 67.2 & 55.0  & 60.2         \\ \hline
			
			\checkmark	&  \checkmark     &     &  &  60.0     &   90.7     & 68.1    &  55.3  &  61.3         \\ \hline
			\checkmark	&       & \checkmark &  &  62.2   &  90.6    &  71.6  &  58.3 &   63.5        \\ \hline
			\checkmark	&    &    &  \checkmark   & 61.9 & 90.3 & 71.0 & 58.0 & 63.3  \\    \hline
			\multicolumn{4}{|c|}{DensePose RCNN* w MID}    & \multicolumn{5}{|c|}{Dense Pose Estimation} \\ \hline
			baseline    &   BBox  & Parts &  keypoints &$AP$   & $AP_{50}$ & $AP_{75}$ & $AP_{M}$ & $AP_{L}$         \\ \hline
			\checkmark	&              &    &     & 64.4  & 90.8  & 73.6 & 60.2  & 65.7          \\ \hline
			
			\checkmark	&  \checkmark     &     &  & 66.0      & \textbf{92.0}        & 74.5    &  62.7  &  67.5         \\ \hline
			\checkmark	&       & \checkmark &     & \textbf{67.6}      &   91.9 & 76.0  & \textbf{64.2}   &   \textbf{69.1}     \\ \hline
			\checkmark	&    &    &  \checkmark   & 67.3 & 91.6 & \textbf{76.8} & 62.7 & 68.8  \\    \hline
		\end{tabular}
	\label{Tab.supl_abstudy_extra_guidance}
	}
\end{table}
\section*{Appendix B: The generalization ability of the MID}
In this section, we conduct an additional experiment to further demonstrate the generalizability of proposed MID. In particular, we investigate the MID on three part segmentation benchmarks, including CIHP\cite{seg_part:cihp}, MHP\cite{humanparsing:understanding} and Fashionpedia\cite{dataset:fashionpedia}, which are most related to the densepose estimation task. We adopt Mask-RCNN\cite{ins_seg:mask_rcnn} as the baseline model. Corresponding experimental results are summarized in Tab.~\ref{Tab:supl_exp_mid_general}. From the results, we observe that the MID improves baseline model by a large margin on all metrics, which is in line with the findings presented in regular manuscript (Sec. IV-C2). The consistent improvements demonstrate the generalizability of proposed method.

\begin{table}[t]
	\caption{Ablation Study. Investigating the effect of MID on three part segmentation benchmarks.}
	\renewcommand\arraystretch{1.2}
	\resizebox{1.0\linewidth}{!}{
		\begin{tabular}{|cc|ccccc|}
			\hline	
			\multicolumn{7}{|c|}{CIHP Part Segmentation} \\ 
			\hline
			Mask-RCNN & MID  & mIoU   & AP$_{30}$ & AP$_{50}$ & AP$_{vol}$ & PCP$_{50}$  \\
			\hline
			\checkmark &            & 48.0 & 88.3  & 44.1 & 46.3  &  46.8  \\
			\checkmark & \checkmark & 54.6 & 91.9  & 62.2 & 53.1  &  60.0  \\
			\hline
			\multicolumn{2}{|c|}{Gains} & \multicolumn{1}{c}{\textbf{+6.6}} & \multicolumn{1}{c}{\textbf{+3.6}} & \multicolumn{1}{c}{\textbf{+18.1}} & \multicolumn{1}{c}{\textbf{+6.8}}  & \multicolumn{1}{c|}{\textbf{+13.2}} \\
			\hline
			\multicolumn{7}{|c|}{MHP Part Segmentation} \\ 
			\hline
			Mask-RCNN & MID  & mIoU   & AP$_{30}$ & AP$_{50}$ & AP$_{vol}$ & PCP$_{50}$  \\
			\hline
			\checkmark &            & 28.7 & 69.1  & 10.6 & 34.0  &  21.5  \\
			\checkmark & \checkmark & 32.7 & 79.2  & 21.5 & 38.9  &  33.2  \\
			\hline
			\multicolumn{2}{|c|}{Gains} & \multicolumn{1}{c}{\textbf{+4.0}} & \multicolumn{1}{c}{\textbf{+10.1}} & \multicolumn{1}{c}{\textbf{+10.9}} & \multicolumn{1}{c}{\textbf{+4.9}}  & \multicolumn{1}{c|}{\textbf{+11.7}} \\
			\hline
			\multicolumn{7}{|c|}{Fashionpedia Part Segmentation} \\ 
			\hline
			Mask-RCNN & MID  & mIoU   & AP$_{30}$ & AP$_{50}$ & AP$_{vol}$ & PCP$_{50}$  \\
			\hline
			\checkmark &            & 28.7 & 70.9  & 12.5 & 35.3  &  18.5  \\
			\checkmark & \checkmark & 32.8 & 85.3  & 24.3 & 41.5  &  30.7  \\
			\hline
			\multicolumn{2}{|c|}{Gains} & \multicolumn{1}{c}{\textbf{+4.1}} & \multicolumn{1}{c}{\textbf{+14.4}} & \multicolumn{1}{c}{\textbf{+11.8}} & \multicolumn{1}{c}{\textbf{+6.2}}  & \multicolumn{1}{c|}{\textbf{+12.2}} \\
			\hline
		\end{tabular}
	}
	\label{Tab:supl_exp_mid_general}
\end{table}

\section*{Appendix C: Comparison with SOTAs on UltraPose}
In this section, we evaluate proposed KTN on a newly released benchmark, i.e., UltraPose\cite{densepose:ultrapose}. Currently, this dataset provides 5K synthesized images, each of which involves only one synthesized person instance. Different from the DensePose-COCO benchmark, 2D-3D correspondences annotations in UltraPose do not have any errors. We evaluate two versions of KTN on this benchmark and report their performance in Tab.~\ref{Tab.comp_ultrapose} for comparison. From the results, we observe that the extended KTN performs the best among all comparison methods, which further demonstrates the effectiveness and generalizability of proposed method. 

\begin{table}[ht]
	\small
	\centering
	\caption{Comparison with the state-of-the-art methods on UltraPose dataset.}
	\label{Tab.comp_ultrapose}
	\renewcommand\arraystretch{1.3}
	\resizebox{1.0\linewidth}{!}{
		\begin{tabular}{|c|c|c|c|c|c|c|}
			\hline
			Method  & $AP$   & $AP_{50}$ & $AP_{55}$ & $AR$   & $AR_{50}$ & $AR_{75}$        \\ \hline
			DensePose R-CNN+uncertainty\cite{densepose:uncertainty}   	&    38.0    & 66.2   &   63.6  & 41.7  & 69.9 & 47.4         \\ 
			DensePose R-CNN \cite{densepose:dphuman}	&  39.6 & 74.3   &   70.0  & 44.4 & 76.1 & 49.9  \\   
			DeepLabV3 Head \cite{densepose:ContinuousSurfaceEmbeddings}	&    43.1   & 75.9    & 74.9      & 47.5        & 78.8    &  55.3 \\ 
			TransUltra \cite{densepose:ultrapose}	&  49.1     & 89.5 &  88.3   &   58.0    &  94.6  & 69.5  \\ 
			KTNv1\cite{densepose:ktn} 	&  46.9    &  93.9 &  91.1   &  53.4    &  96.0  &  60.0 \\ \hline
			KTNv2 (Ours)	&  \textbf{51.8}     & \textbf{94.6} &  \textbf{91.3}   &   \textbf{59.0}    &  \textbf{97.0}  & \textbf{73.5}  \\ \hline
		\end{tabular}
	}
\end{table}

\section*{Appendix D: The bottleneck of DensePose models}
The performance of densepose algorithm is jointly decided by multiple sub-tasks, including body segmentation, surface classification and UV coordinate regression. In this section, we gauge the relative difficulty of these sub-tasks, namely, which sub-task is the main bottleneck of the proposed KTN. We choose the KTN with ResNet50 backbone as the baseline and evaluate it on the DensePose-COCO \textit{minival} set. During the evaluation, we simply replace predictions with corresponding ground truth when investigating each sub-task.

The quantitative results are shown in Tab.~\ref{Tab.abstudy_gt}, where we can see that using ground truth body masks improves AP from 68.3\% to 72.4\%. It can be further improved by 0.3\% and 3.3\% (row 3 and row 4) when replacing predicted UV coordinates and 3D surface part categories with ground truth, respectively, which indicates that our KTN can make accurate prediction for UV coordinates when corresponding points are correctly classified. After replacing predictions of 3D surface with ground truth (row 4), we gradually replace V coordinates and U coordinates with their corresponding ground truth, as shown in row 5, 6 and 7.
From the results, we can observe that replacing all 3D predictions with ground truth brings large improvements (19.7\%), where major improvements come from substitution of UV coordinates (16.4\%).
It indicates that the UV regression is the main bottleneck in densepose estimation, where the regression of V coordinates is main limitation (see row 5 and 6).
\begin{table}[ht]
	\small
	\setlength{\belowcaptionskip}{-5pt}%
	\setlength{\abovecaptionskip}{5pt}%
	\caption{Ablation study. Investigating the bottleneck of the proposed method. $G_b$, $G_{sp}$, $G_{v}$ and $G_{u}$ respectively denotes ground truth body mask, ground truth 3D surface mask, ground truth V coordinates and ground truth U coordinates.}
	\label{Tab.abstudy_gt}
	\resizebox{\linewidth}{!}{
		\begin{tabular}{|ccccc|ccc|}
			\hline
			\multicolumn{5}{|c|}{Setting}    & \multicolumn{3}{|c|}{Dense Pose Estimation} \\ \hline
			KTN-net & $G_b$ & $G_{sp}$ & $G_{v}$ & $G_{u}$ & $AP$          & $AP_{50}$         & $AP_{75}$         \\ \hline
			\checkmark	&        &    &    &    &     68.3\%        &     92.1\%         &     77.4\%         \\ \hline
			\checkmark	&   \checkmark     &    &    &    &    72.4\%         &    92.9\%          &    82.7\%          \\ \hline
			\checkmark	&   \checkmark     &    &  \checkmark  &   \checkmark &        72.7\%     &      93.1\%        &   83.1\%          \\ \hline
			\checkmark	&   \checkmark    &  \checkmark  &    &    &        75.7\%     &      94.2\%        &   91.2\%           \\\hline
			\checkmark	&   \checkmark     & \checkmark   & \checkmark   &    &        80.1\%     &     94.3\%         &  92.2\%            \\ \hline
			\checkmark	&   \checkmark     & \checkmark   &    &   \checkmark &       89.4\%     &      94.9\%        &   93.8\%           \\ \hline
			\checkmark	&   \checkmark     & \checkmark   &  \checkmark  &   \checkmark &        92.1\%     &     94.9\%         &  93.8\%           \\ \hline
		\end{tabular}
	}
\end{table}
\begin{table}[t]
	\centering
	\caption{\# Params and GFLOPs of KTN. The FLOPs is computed with one proposal. M=$10^6$,G=$2^{30}$}
	\label{Tab.modelsize}
	\renewcommand\arraystretch{1.2}
	\resizebox{0.99\linewidth}{!}{
		\begin{tabular}{|c|c|c|p{1cm}p{1cm}|c|}
			\hline
			\multicolumn{6}{|c|}{Inference} \\ \hline
			\multirow{2}{*}{Backbone}   & \multirow{2}{*}{\textbf{\#}Params }      & \multirow{2}{*}{FLOPs}  &   \multicolumn{2}{|c|}{Speed per image}  & \multirow{2}{*}{mAP} \\ \cline{4-5}
			&      &        &  \multicolumn{1}{|c|}{\textit{forward}}  &  \multicolumn{1}{|c|}{\textit{post-process}}    &     \\\hline                       
			ResNet50 &  49.4M                 & 20.0G                    &     \multicolumn{1}{|c|}{0.169 s/img}      &    \multicolumn{1}{|c|}{0.479 s/img}     &        68.3                           \\ \hline
			ResNeXt101 &  117.5M                 & 33.0G                    &   \multicolumn{1}{|c|}{0.222 s/img}     &      \multicolumn{1}{|c|}{0.426 s/img}     &      69.2                                    \\ \hline
			HRNetW32 &  55.2M                     &   200.5G                   &  \multicolumn{1}{|c|}{0.242 s/img}   &  \multicolumn{1}{|c|}{0.443 s/img}     &          71.8                       \\ \hline	
			HRNetW48 &  91.2M                 & 362.7G                    &      \multicolumn{1}{|c|}{0.253 s/img}    &   \multicolumn{1}{|c|}{0.408 s/img}   &           71.9                                  \\ \hline
			\hline
			\multicolumn{6}{|c|}{Training} \\ \hline
			\multicolumn{1}{|c|}{Backbone}   & \multicolumn{1}{|c|}{Device}  & \multicolumn{1}{|c|}{Mem}    & \multicolumn{1}{|c|}{Batch size}  &   \multicolumn{1}{|c|}{Iterations}  & \multicolumn{1}{|c|}{Training Costs} \\ \hline
			ResNet50 &  Titan X                & 12G                    &     \multicolumn{1}{|c|}{8}      &    \multicolumn{1}{|c|}{260k}    &              54 hours                   \\ \hline
			ResNet50 &  Titan RTX              & 24G                    &     \multicolumn{1}{|c|}{12}  &     \multicolumn{1}{|c|}{170k}  &               28 hours                 \\ \hline
			ResNet50 &  Tesla V100             & 32G                    &     \multicolumn{1}{|c|}{16}      &    \multicolumn{1}{|c|}{130k}    &              21 hours                   \\ \hline
			HRNetW32 &  Titan X                & 12G                    &     \multicolumn{1}{|c|}{4}   &      \multicolumn{1}{|c|}{520k}     &           260 hours                             \\ \hline
			HRNetW32 &  Titan RTX              & 24G                    &     \multicolumn{1}{|c|}{12}  &     \multicolumn{1}{|c|}{170k}  &               176 hours                 \\ \hline	
			HRNetW32 &  Tesla V100             & 32G                    &     \multicolumn{1}{|c|}{16}  &   \multicolumn{1}{|c|}{130k}       &          46 hours                          \\ \hline
		\end{tabular}
	}
\end{table}
\section*{Appendix E: Efficiency of DensePose models}
In this section, we provide computation efficiency comparison among KTN with different backbones. Specifically, we respectively provide timing costs at two phases: inference and training. The comparison results are summarized in Tab.\ref{Tab.modelsize}. 

For inference phase, we record the model size, GFLOPs, speed and densepose estimation results, respectively. Specifically, the inference time is counted with single-scale ($800\times1333$) testing on one Titan RTX GPU, and the GFLOPs metric is computed with one proposal. From the Tab.\ref{Tab.modelsize}, we can see that the parameter size of the KTN with ResNet50 backbone is around 49.4M, which is the smallest model. In terms of computational complexity, the KTN with HRNetW48 requires the most, which is 362.7 GFLOPs. As for speed comparison, all models process around 1.2 frames per second (fps). In particular, ResNet based models have lower timing costs in network forwarding, while requiring more costs on post-process. However, the KTN with HRNet as backbone present the opposite result. When replacing HRNetW32 with HRNetW48, the performance of densepose estimation slightly increases, but the computation cost dramatically increases. This suggests that the performance gain obtained by simply extending network capacity already reaches the bottleneck.

As for the training phase, we compare pyramidal model (ResNet) with front-to-end high-resolution model (i.e., HRNet). In particular, the setting of batch size is limited by the device. Here, we choose three devices that commonly used for training deep learning models: 1) Titan X with 12G memory. 2) Titan RTX with 24G memory; and 3) Tesla V100 with 32G memory. Furthermore, each model is trained on four GPU cards. From the results reported in Tab.\ref{Tab.modelsize}, we can find that training front-to-end high-resolution model requires much large training costs on Titan X (i.e., 260 hours), compared with pyramidal model (21 hours). When using more powerful devices (i.e., Titan RTX or Tesla V100), training costs for hrnet decreases dramatically, but still larger than the training costs for ResNet. This is why we follow the design rule of pyramidal structure, where it shows the characteristic of device friendly.

\section*{Appendix F: More Qualitative Analysis}

\noindent\textbf{Qualitative Analysis:} We visualize prediction results of KTN with ResNet50 backbone for real-world scenes with different detection complexity:
1) Easy: An image contains few persons or multiple persons, where the size of each person is large enough for detection, and they are neatly arranged without occlusion or with slight occlusion;
2) Hard: One image involves crowds with heavy overlapping or contains some humanoid objects which may be misclassified as a person. 

\begin{figure*}[ht]
	\setlength{\abovecaptionskip}{-0.1cm}%
	\setlength{\belowcaptionskip}{-0.3cm}%
	\begin{center}
		\includegraphics[width=1.0\linewidth,height=0.9\linewidth]{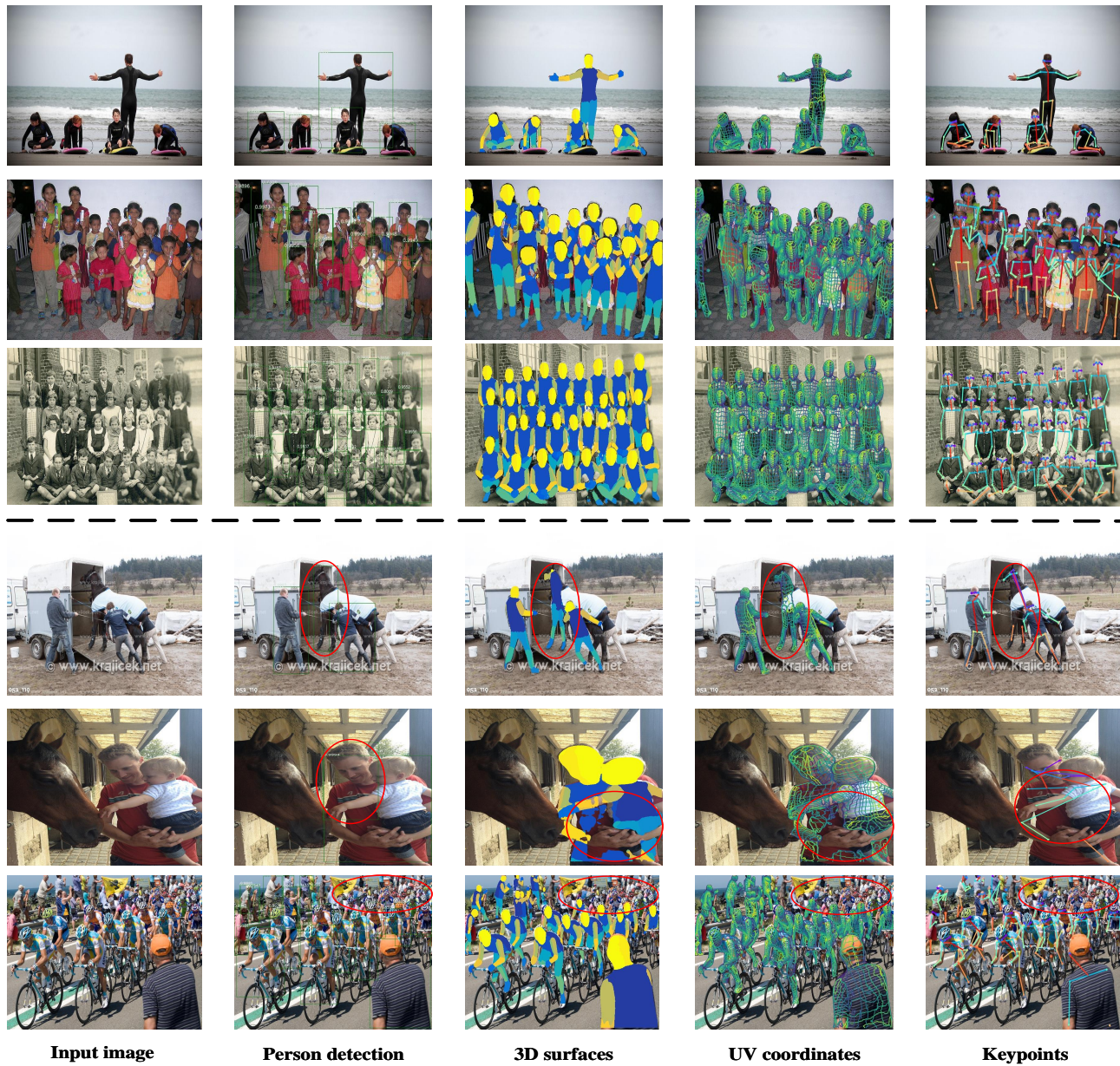}
	\end{center}
	\caption{Visualization results predicted from KTN. From left to right: input image, bounding boxes, 3D surfaces, UV coordinates and keypoints detection. The red circles spot the failure cases of KTN.}
	\label{fig:qua_dpcom}
\end{figure*}

The visualization results are summarized in Fig.~\ref{fig:qua_dpcom}. As can be seen, the KTN can produce accurate 3D surface segmentations and smooth contours of 3D body surface under the condition of ``in the wild'', where the input image involves persons with various postures or multi-person with light occlusion. However, the KTN produces unsatisfactory results, such as a large area of false positives (row 4) and missing estimations (row 5 and row 6), especially when it comes to the condition of ``heavy occlusion''. There are two reasons behind that: 1) Incorrect detection results. The two-stage pipeline assumes that each predicted bounding box contains one person instance. Therefore, densepose estimation system will be confused by detection results with low quality. 2) Severe occlusion. There are few densepose annotations of occluded person in DensePose-COCO dataset, where densepose estimation systems, including but not limited to KTN, are trained by general person instances without occlusion modeling. 

\begin{figure*}[ht]
	\centering
	\setlength{\abovecaptionskip}{-0.1cm}%
	\begin{center}
		\includegraphics[width=0.99\linewidth,height=1.1\linewidth]{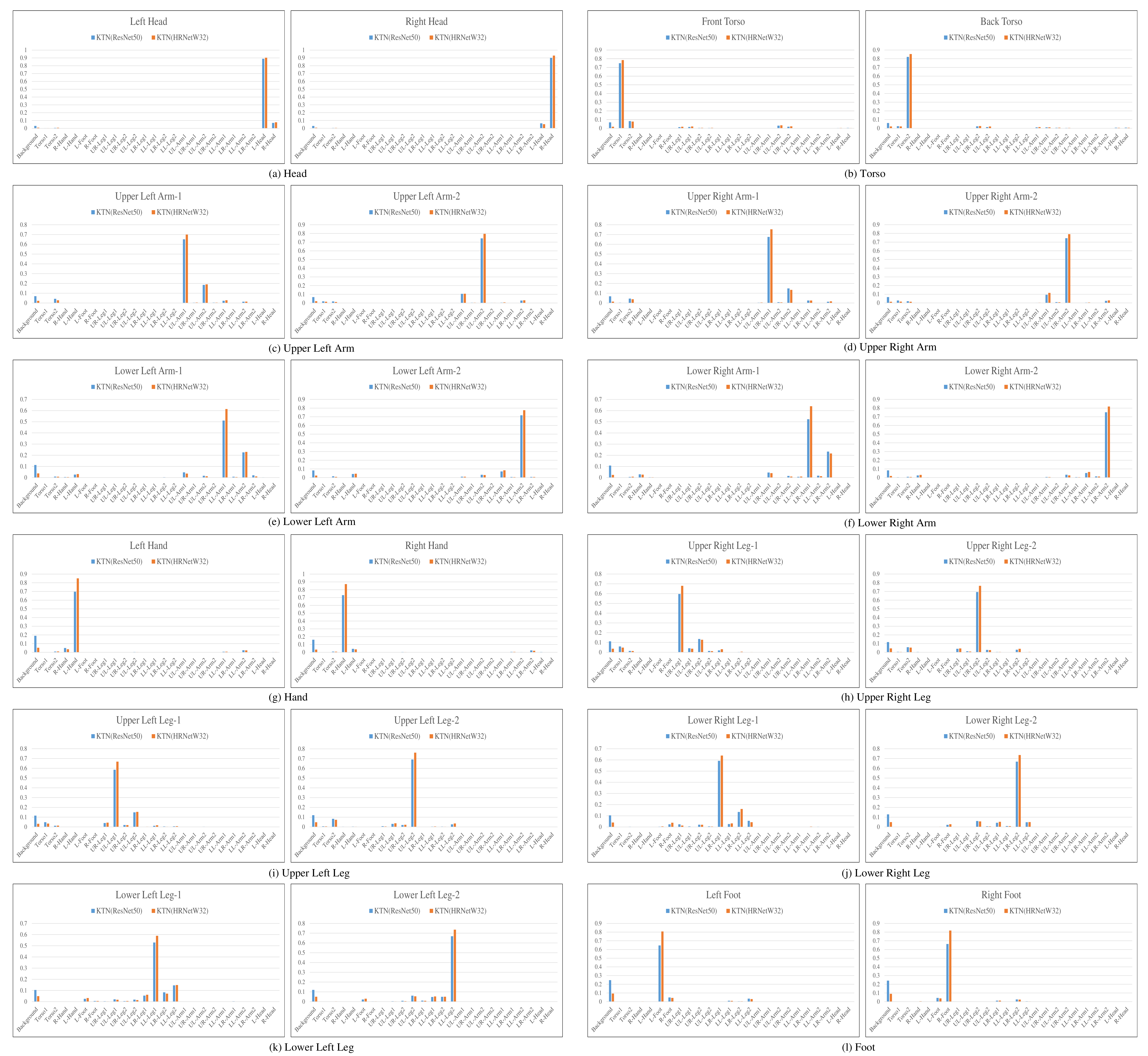}
	\end{center}
	\centering
	\caption{Prediction distributions of each surface category, involving 10 standard body parts with their corresponding sub-surfaces. Each sub-figure presents the statistics of point-wise prediction for specific surface, where predictions are obtained from KTN with ResNet50 backbone and HRNetW32 backbone, respectively. Zoom in for a better view.}
	\label{fig:prediction_distribution}
\end{figure*} 

\noindent\textbf{Errors Analysis:} In terms of surface classification, we perform an additional point-wise evaluation on the predictions obtained from KTN with different backbones (i.e., ResNet-50 and HRNet-w32) and assess them by visualizing prediction distributions of each category, which are summarized in Fig.~\ref{fig:prediction_distribution}. From the quantitative visualization, we observe that there are two main deficiencies that existed in our densepose model. First, most failure cases suffer from misclassifying distinguishing adjacent surfaces. For example, model has trouble distinguishing up left arm-1 from up left arm-2 (Fig.~\ref{fig:prediction_distribution}(e)). Similar issue also can be observed in leg related prediction results (see Fig.~\ref{fig:prediction_distribution}(h-k)). Second, the KTN models lack the ability to distinguish between left and right. As shown in Fig.~\ref{fig:prediction_distribution}. (g), some samples (around 8\%) of left head are recognized as right head, and some samples of left hand (around 5\%) are misclassified as right hand. There are two possible reasons behind such left-right misclassification: 1) The annotation resources in the current DensePose-COCO dataset are too sparse to train a powerful densepose system; and 2) Highly similar appearances provide few clues for fine-grained discrimination, which is even difficult to be captured by human beings.
\end{document}